\documentclass[10pt,twocolumn,letterpaper]{article}

\usepackage{epsfig,times}
\usepackage{graphicx}
\usepackage{amsmath}
\usepackage{amssymb}

\usepackage{threeparttable}
\usepackage[vlined, ruled, linesnumbered]{algorithm2e}
\usepackage{multirow}
\usepackage{booktabs}
\usepackage{bbm}
\usepackage{pifont}
\usepackage{subfig}

\usepackage{cite,xspace}
\usepackage[margin=0.65in]{geometry}

\def\etal{{\em et al.}\xspace}
\usepackage[pagebackref=true,breaklinks=true,letterpaper=true,colorlinks,bookmarks=false]{hyperref}

\usepackage{caption}
\begin{document}

\title{\Large
{\bf
DiverseDepth: Affine-invariant Depth Prediction Using Diverse
Data
}
\\[.5cm]
\large
        Wei Yin$ ^1 $ ~ ~ ~
        Xinlong Wang$ ^1 $ ~ ~ ~
        Chunhua Shen$ ^1 $\thanks{Corresponding author. E-mail: $\sf chunhua.shen@adelaide.edu.au $ } ~ ~
        Yifan Liu$ ^1 $ ~ ~ ~
        Zhi Tian$ ^1 $ ~ ~ ~
        \\
        Songcen Xu$ ^2 $ ~ ~ ~
        Changming Sun$ ^3 $ ~ ~ ~
        Dou Renyin$ ^4 $
\\[0.3cm]
        $^1$ University of Adelaide ~ ~
        $^2$ Huawei Noah’s Ark Lab ~ ~
        $^3$ Data61 ~ ~
        $^4$ Beijing University of Posts \& Telecommunications
}

\date{}

\maketitle

\newcommand{\datasetfullname}{Diverse Scene Depth dataset (DiverseDepth)}
\newcommand{\datasetshortname}{DiverseDepth\xspace}

\begin{abstract}
We present a method for depth estimation with monocular images,
which can predict
high-quality depth on diverse scenes
up to an affine transformation, thus preserving accurate shapes of a scene.
Previous
 methods  that predict metric depth
 often
 work well only for %
 a specific scene.
 In contrast, learning %
 relative depth (information of being closer or further) %
 can enjoy better %
 generalization,
 with the price of failing to recover the accurate geometric shape of the scene.

 In this work,
 we propose a dataset and methods to tackle this dilemma, aiming to predict accurate
 depth up to an affine transformation with good generalization to diverse scenes.
 First we construct a large-scale and diverse dataset, termed \datasetfullname, which has a broad range of scenes and foreground contents. Compared with previous
 learning
 objectives, i.e.,  learning metric depth or relative depth, we propose to learn the affine-invariant depth
 using
 our diverse dataset to ensure both
 generalization
 and high-quality geometric %
 shapes of scenes.
 Furthermore, in order to train the model on the complex dataset
 effectively, we propose a multi-curriculum  learning method.
 Experiments
show
 that our method outperforms previous methods on $8$
 datasets by a large margin
 with the zero-shot test setting, demonstrating the excellent generalization capacity of the learned model to diverse scenes.
 The reconstructed point clouds with the predicted depth show that our method can recover high-quality 3D shapes. Code and dataset are available at:
 \url{https://tinyurl.com/DiverseDepth}

\end{abstract}

\section{Introduction}
Monocular depth estimation    %
is a challenging problem. As  %
there exists no easy way to enforce
geometric constraints %
to recover the depth from a still image, various data-driven approaches are proposed to
exploit
comprehensive cues~\cite{fu2018deep, Yin2019enforcing, dai2017scannet,Cao2017}.

\begin{figure}[!bth]
\centering
\includegraphics[width=0.47\textwidth]{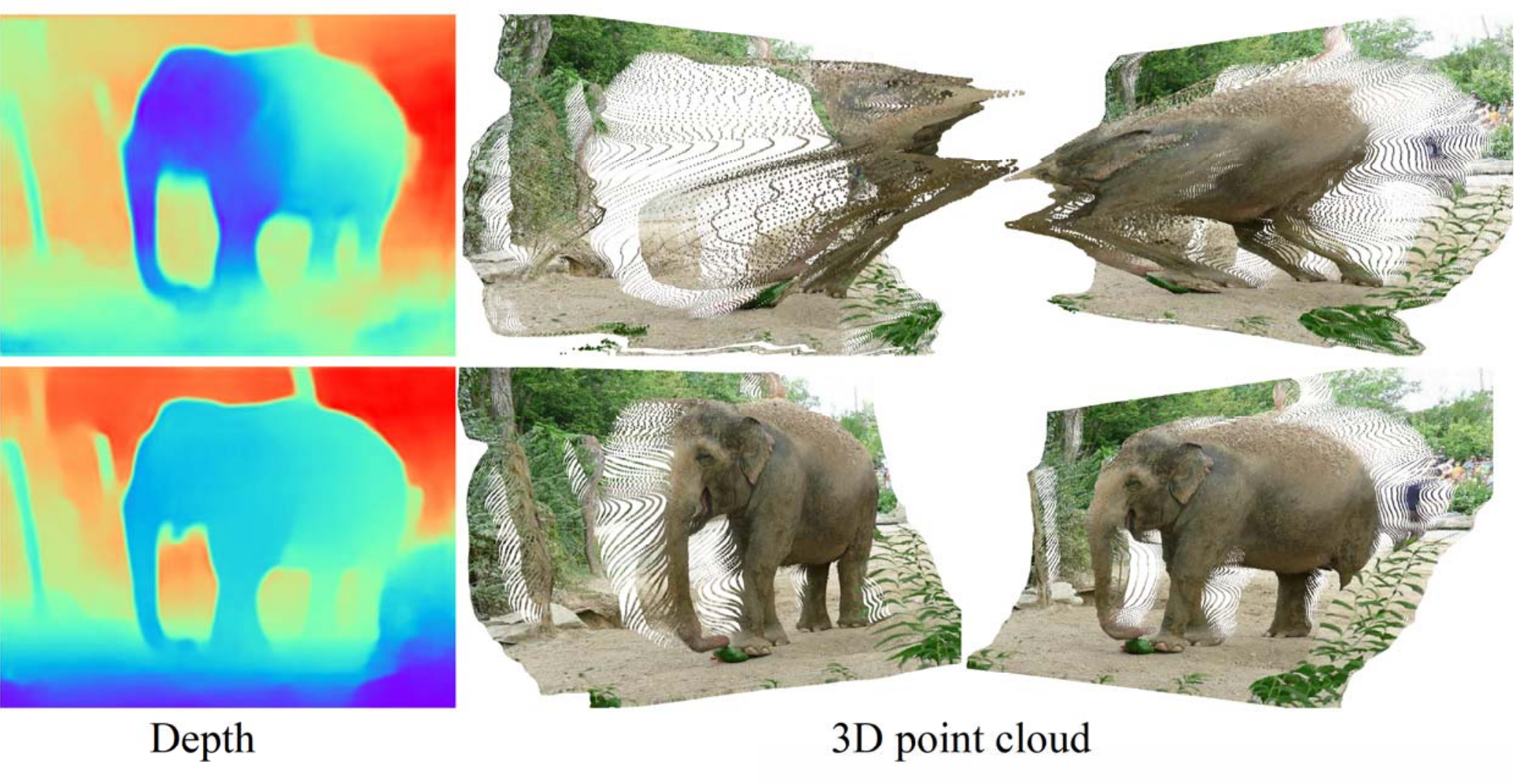}
\caption{Qualitative comparison of depth and reconstructed 3D point cloud between our method and
that of the recent
learning relative depth method of Xian~\etal~\cite{xian2018monocular}. The first row is the predicted depth and reconstructed 3D point cloud from the depth of theirs, while the second row is ours.
The relative depth model fails to
recover the 3D geometric shape of the scene (see the distorted elephant and ground area).
Ours does much better.
 Note that
this test image is sampled from
the
DIW dataset, which does not overlap with our training data. }
\label{fig:intro cmp}
\vspace{-1em}
\end{figure}

Previous %
methods of depth estimation based on deep convolutional neural networks (DCNN) have achieved outstanding performance on popular benchmarks~\cite{Yin2019enforcing, fu2018deep, alhashim2018high}. They can be mainly summarized into two categories. (1) The first group enforces the pixel-wise metric supervision
to produce the accurate metric depth map
typically on some
specific scenes, such as indoor environments, but
in general does not
work well on diverse scenes. For example,
the recent virtual normal method of
\cite{Yin2019enforcing} can achieve state-of-the-art performances on various benchmarks with the training and testing done on each benchmark separately.
(2) The second group aims to address the
issue of
generalization to multiple scene data
by learning with relative depth such that large-scale datasets of diverse scenes
can be collected much easier. A typical example is the depth-in-the-wild (DIW) dataset~\cite{chen2016single}.
Such methods often explore the pair-wise ordinal relations for learning and
only the relative depth can be predicted.
A clear drawback is that these models fail to
recover the high-quality geometric
3D shapes as only ordinal relations are used in learning.
For example, the reconstructed 3D point cloud from the relative depth (first row in Figure~\ref{fig:intro cmp}) is completely distorted and cannot represent the shape of the elephant.
In order to ensure both
good generalization
and high-quality 3D
depth information,
there are two obstacles: (1) lacking
diverse and high-quality training data; (2)
an appropriate
learning objective function that is easy to optimize,
yet preserving as much geometric information as possible.

In this work, we seek to address these problems from three aspects: (1) constructing a large-scale dataset with diverse scenes, \datasetfullname, including both rigid and non-rigid contents in both indoor and outdoor environments. With our proposed dataset construction method, such a dataset can be %
relatively easy to expand.
Existing datasets are either  difficult to expand (metric depth), or
only annotated with weak geometric information (relative depth).
Our dataset strikes a balance between these.
%
%
(2) enforcing the DCNN model to learn the affine-invariant depth instead of a specific depth value or relative depth on the diverse scales dataset; (3) proposing a multi-curriculum learning method for the effective training on this complex dataset.

Current available
RGB-D datasets can be summarized into two categories: (1) RGB-depth pairs captured by
a
depth sensor have high precision,
typically %
accommodating
only few scenes as it can be very costly to acquire a very large dataset of diverse scenes.
For example, the KITTI dataset \cite{geiger2013vision} is captured %
with
LIDAR on road scenes only, while the NYU dataset \cite{silberman2012indoor} only contains several indoor rooms. (2) Images with much more diverse scenes are
available
online and can be annotated with coarse
depth with reasonable effort.  The large-scale DIW~\cite{chen2016single} dataset is manually annotated with only one pair of ordinal depth relations for each image.
To construct our large and diverse dataset, we harvest
stereoscopic videos and images with diverse contents and use %
stereo matching methods to
obtain depth maps.
The dataset
contains
both rigid and non-rigid foregrounds, such as people, animals, and cars.
Our \datasetshortname is more diverse than metric depth datasets, while it contains more geometric information than existing relative depth datasets because depth in our dataset is metric depth up to an affine transformation.
We have sampled some images from Taskonomy~\cite{zamir2018taskonomy} and DIML~\cite{cho2019large} and
added them into our dataset.

The commonly used learning objectives can be summarized into two categories: (1) directly minimizing the pixel-wise divergence to the ground-truth metric depth~\cite{fu2018deep, eigen2014depth, Yin2019enforcing}; (2) exploiting the uniformity of pair-wise ordinal relations~\cite{chen2016single, xian2018monocular}.
However, both two methods cannot well balance the high generalization and enriching the model with abundant geometric information. In contrast, we reduce the difficulty of depth prediction by explicitly disentangling depth scales during training. The model will ignore the depth scales and make the predicted depth invariant to the affine transformation, (i.e., translation, scale). Several loss functions can satisfy the requirement. For example, the surface normal and virtual normal loss~\cite{Yin2019enforcing} are affine-invariant because they are based on normals. Besides, the scale-and-shift-invariant loss (SSIL)~\cite{lasinger2019towards} explicitly recovers the scaling and shifting gap between the predicted and ground-truth depth.  We combine a geometric constraint and SSIL to supervise the model. The second row in Figure~\ref{fig:intro cmp} shows the predicted affine-invariant depth of a DIW image and the reconstructed 3D point cloud, which can clearly represent the shape of the elephant and ground. Experiments on $8$ zero-shot datasets show the effectiveness of learning affine-invariant depth.

Furthermore, training the model on the large-scale and diverse dataset effectively is also a problem. We propose a multi-curriculum learning method for training. Hacohen and Weinshall~\cite{hacohen2019power} have proved that an easy-to-hard curriculum will not change the global minimum of the optimization function but increase the learning speed and improve the final performance on test data. Here, we separate the diverse learning materials to different curriculums and introduce each curriculum with increasing difficulty to the network. Experiments show this method can significantly promote the performance on various scenes.

In conclusion, our contributions are outlined as follows.
\begin{itemize}
    \itemsep -0.125cm

\item We construct a large scale and high-diversity RGB-D dataset, \datasetshortname;

\item We are the first to propose to learn affine-invariant depth on the diverse dataset, which ensures both high generalization and high-quality geometric shapes of scenes;

\item We propose a multi-curriculum learning method to effectively train the model on the large-scale and diverse dataset. Experiments on $8$ zero-shot datasets show our method outperforms previous methods noticeably.

\end{itemize}

\subsection{Related Work}
\noindent\textbf{Monocular depth estimation.}
It is an important task for numerous applications.
Deep learning methods have %
dramatically advanced
many
computer vision tasks, %
 including depth estimation.
Existing
methods may be categorized into supervised learning~\cite{Yin2019enforcing, xian2018monocular, fu2018deep} and unsupervised/self-supervised learning methods~\cite{bian2019depth, zhou2017unsupervised, godard2019digging}. Eigen \etal \cite{eigen2014depth} propose the first multi-scale network for this task.
Since then
various
powerful network architectures~\cite{fu2018deep} are proposed.
Unsupervised learning of depth is also very popular recently because no ground-truth depths are needed for the system. Zhou~\etal~\cite{zhou2017unsupervised} are the first ones to demonstrate an approach to jointly predict the depth and the ego-motion from the monocular video. Then various techniques~\cite{Yin2019enforcing, cheng2019learning, jiao2018look} are developed to promote the performance. However, the key drawback of such methods is that they often cannot well generalize to unseen scenes outside the training set.
We propose a dataset and training methods to tackle this   generalization issue
of depth estimation.

\noindent\textbf{RGB-D datasets.} Datasets~\cite{saxena2008make3d, geiger2013vision, silberman2012indoor, dai2017scannet} are significant for the advancement of data-driven depth prediction methods. KITTI~\cite{geiger2013vision} and NYU~\cite{silberman2012indoor} are the most popular datasets, captured by LIDAR on outdoor streets and Kinect in indoor rooms. To further promote the performance of %
depth estimation
and study the relationships between different vision tasks, more larger sized RGB-D datasets are constructed, such as ScanNet~\cite{dai2017scannet}, Taskonomy~\cite{zamir2018taskonomy}. However, such datasets only contain very few scenes. By contrast, to solve the generalization of depth estimation
methods on diverse scenes, Chen \etal~\cite{chen2016single} propose the DIW dataset, Xian \etal \cite{xian2018monocular} construct the RedWeb dataset, and Li and Snavely~\cite{li2018megadepth} release the MegaDepth dataset.

\noindent\textbf{Curriculum learning.}
In many applications, introducing concepts in ascending difficulty to the learner is a common practice.
Several works have demonstrated that curriculum learning~\cite{weinshall2018curriculum, hacohen2019power, bengio2009curriculum} can
boost the performance of deep learning methods. Weinshall~\etal~\cite{weinshall2018curriculum} combine the transfer learning and curriculum learning methods to construct a better curriculum, which can improve both the speed of convergence and the final accuracy. Besides, Hacohen and Weinshall~\cite{hacohen2019power} propose a bootstrapping method to train the network by self-tutoring.

\section{Approach}

\begin{table}[]
\centering
\caption{Comparison of previous RGB-D datasets.}
\scalebox{0.8}{
\begin{tabular}{l|llll}
\toprule[1pt]
Dataset   & Diversity & Dense & Accuracy & Images \\ \hline\hline
\multicolumn{5}{c}{Captured by RGB-D sensor} \\ \hline \hline
NYU~\cite{silberman2012indoor}& Low    & \checkmark      & High    & $407$K       \\
KITTI~\cite{geiger2013vision} & Low      & \checkmark      & High    & $93$K       \\
SUN-RGBD~\cite{song2015sun}  & Low      & \checkmark      & High    & $10$K       \\
ScanNet~\cite{dai2017scannet}& Low      & \checkmark      & High    & $2.5$M       \\
Make3D~\cite{saxena2008make3d}& Low      & \checkmark      & High  & $534$       \\\hline\hline
\multicolumn{5}{c}{Crawled online} \\ \hline \hline
DIW~\cite{chen2016single}  & High     &           & Low     & $496$K        \\
Youtube3D~\cite{chen2019learning}& High     &           & Low     & $794$K        \\
RedWeb~\cite{xian2018monocular}& Medium & \checkmark      & Medium  & $3.6$K    \\
MegaDepth~\cite{li2018megadepth}& Medium   & \checkmark      & Medium  & $130$K       \\\hline\hline
Ours      & High     & \checkmark      & Medium  & $320$K      \\ \toprule[1pt]
\end{tabular}}
\vspace{-1.5 em}
\label{table:datasets}
\end{table}

\subsection{Diverse Scene Depth Dataset Construction}
\noindent\textbf{Dataset statistics.} Table~\ref{table:datasets} compares the released popular RGB-D datasets. RGB-D sensors can
capture high-precision depth data, but they only contain limited scenes. By contrast, crawling large-scale online images can promote scene diversity.
Previous datasets only have sparse ordinal depth annotations, such as DIW and Youtube3D. Although RedWeb and MegaDepth advance the ground-truth depth quality, RedWeb only has 3600 images and MegaDepth only contains static scenes.

Therefore, to feature diversity, quality, as well as data size, we construct a new dataset with multiple data collection sources. Firstly, we collect large-scale online stereo images and videos to construct foreground objects part, termed \textit{Part-fore}, such as plants, people and animals. Then, we sample some images from Taskonomy and DIML to constitute the indoor and outdoor background part, termed \textit{Part-in} and \textit{Part-out}.

\noindent\textbf{Foreground part.} %
The processes of \textit{Part-fore} data construction is outlined as follows.

(1) Crawling online stereoscopic images and videos.
We summarize %
three websites for data collection: Flickr, 3DStreaming and YouTube. We firstly discard invalid frames and images that are not left/right stereo contents by comparing the similarity of left/right parts, i.e., removing low similarity frames. Then we manually inspect outliers.
%
%
%
%
%
%

(2) Retrieving disparities from stereo materials, then reversing and scaling them to obtain depths. As parameters of all stereo cameras are unknown, we cannot rectify the stereo images, remove lens distortion, and align the epipolar line. Existing stereo matching methods~\cite{hirschmuller2007stereo} based on comparing the local or semi-global features along the epipolar line cannot obtain the disparity. Instead, we utilize the optical flow~\cite{ilg2017flownet} method to match the paired pixels in stereo samples and take the horizontal matching as the disparity. The depth is obtained by reversing and scaling the disparity.

\begin{figure*}[!t]
\centering
\includegraphics[width=0.9775\textwidth]{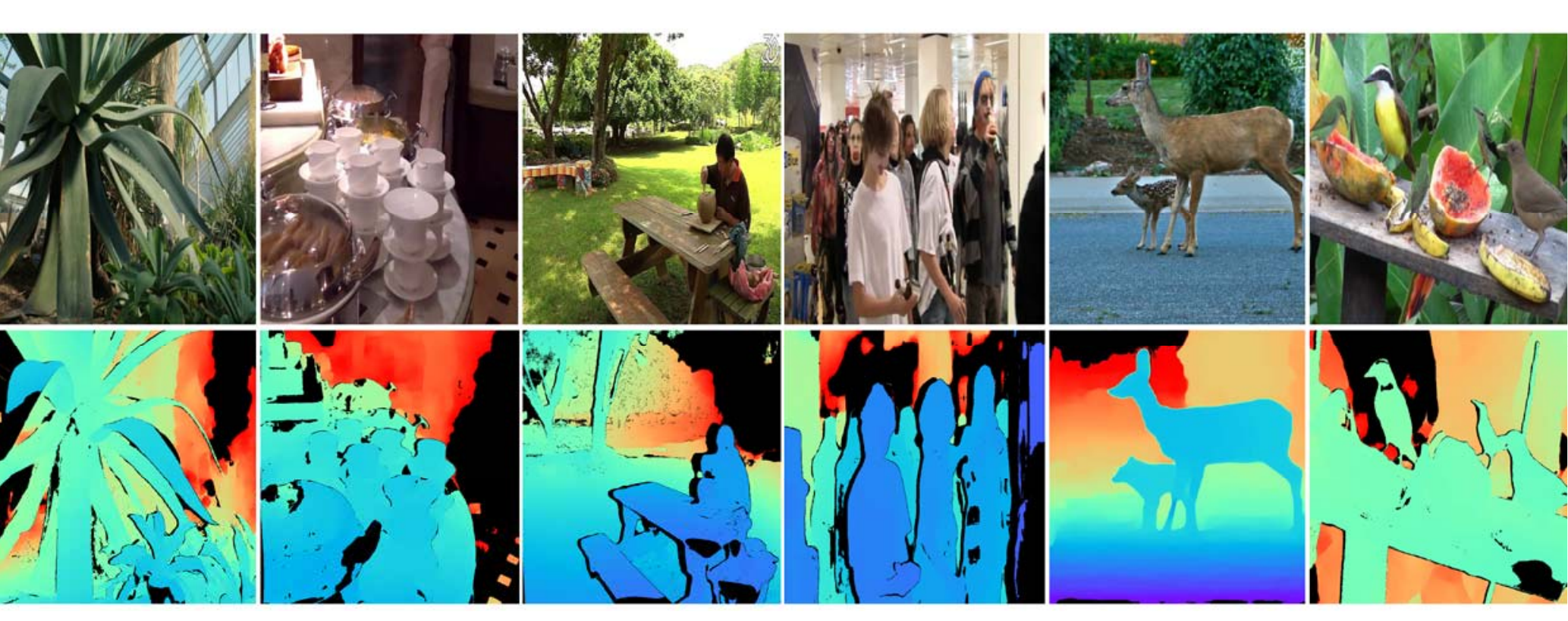}
\caption{Examples of the \datasetshortname dataset. Purple parts are closer, while red regions are farther.}
\label{fig:dataset examples}
\end{figure*}

(3) Filtering depth maps. We find that many outliers and noises residing in depths are mainly caused by large distortions, small baselines, and poor images features. Here, we take $3$ metrics to mask out such noises. Firstly, pixels with vertical disparities larger than $5$ are removed. Secondly, pixels with the left-right disparity difference greater than $2$ are removed. Furthermore, images with valid pixels less than $30\%$ are discarded. After these filtering processes, We totally collect more than $90$K RGB-D pairs for the \textit{Part-fore}. Example images are shown in Figure~\ref{fig:dataset examples}.

\noindent\textbf{Background part.} In order to enrich the diverse background environments, we sample $100$K images from an indoor and an outdoor dataset respectively, i.e.,  Taskonomy~\cite{zamir2018taskonomy} and DIML~\cite{cho2019large}. The Taskonomy samples constitute our indoor background data, \textit{Part-in}, while the DIML ones are the outdoor background part, \textit{Part-out}.

Therefore, our \datasetshortname dataset has around $300$K diverse RGB-D pairs, which is composed of three different parts, i.e., \textit{Part-fore}, \textit{Part-in} and \textit{Part-out}. There are around $18$K images for testing.

\subsection{Affine-invariant Depth Prediction}
\begin{figure}[!bt]
\centering
\includegraphics[width=0.4\textwidth]{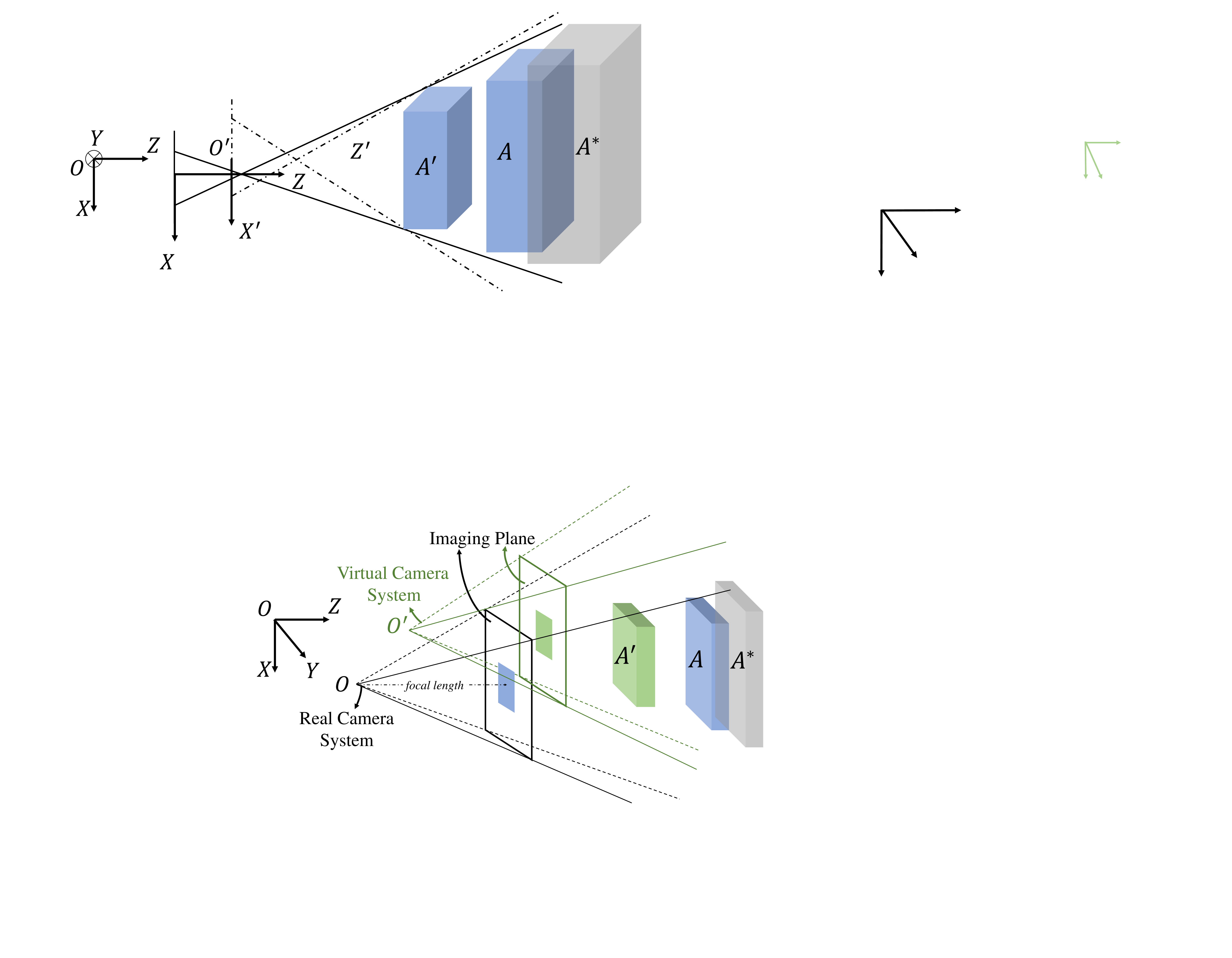}
\caption{The geometric model of an imaging system. $A^{\ast}$ is the ground-truth location for an object. $A$ is the predicted location by learning metric depth method, while $A{'}$ is the predicted location by our learning affine-invariant depth method. }
\label{fig:geometric model}
\vspace{-1.5em}
\end{figure}

The geometric model of the monocular depth estimation system is illustrated in Figure~\ref{fig:geometric model}.  The ground-truth object in the scene is $A^{\ast}$, and the real camera system is $O$-$XYZ$ (the black one in Figure~\ref{fig:geometric model}). When learning the metric depth, the model $\mathcal{G}(\bf{I}, \theta)$ may predict the object at location $A$. $\bf{I}$ is the input image. The learning objective of such methods is to minimize the divergence between $A$ and $A^{\ast}$, i.e.,  $ \min_{\theta}\left | \mathcal{G}({\bf I}, \theta) - d^{\ast} \right |$, where $d^{\ast}$ is the ground-truth depth and $\theta$ is the network parameters. As such methods mainly train and test the model on the same benchmark, where the camera system and the scale remain almost the same, the model can implicitly learn the camera system and produce accurate depth on the testing data~\cite{dijk2019neural}. The typical loss functions for learning metric depth are illustrated in Table~\ref{table:loss_list}. However, when training and testing on diverse dataset, where the camera system and scale vary,
it is  theoretically  not possible for the model to accommodate multiple camera parameters.
The tractable approach is to feed camera parameters of different camera systems
to the network as part of the input in order to predict metric depth. This requires the access to camera parameters, which are often not available when harvesting online image data.
Our experiments
show
failure cases
of learning metric depth on the diverse dataset (see Table~\ref{table: zero-shot comparison}, Table~\ref{table: different constraints}, and Figure~\ref{fig:overall cmp}).

Learning the relative depth reduces the difficulty of depth prediction from predicting the accurate metric depth to the ordinal relations. With enough diverse training data, this method can predict relative depth on diverse scenes, but it loses geometric information of the scene, such as the geometric shape. For example, the reconstructed 3D point cloud from the relative depth in Figure~\ref{fig:pcd cmp} and Figure~\ref{fig:intro cmp} cannot represent the shape of the sofa and elephant respectively.

In this paper, we propose to learn the affine-invariant depth from the diverse dataset. On the diverse dataset, we define a virtual camera system, $O'$-$X'Y'Z'$ (the green one in Figure~\ref{fig:geometric model}), which has the same viewpoint as the real one but has the different optical center location and the focal length. Therefore, there is an affine transformation, i.e.,  translation $T$ and scaling $s$, between the real camera system $O$-$XYZ$ and the virtual one $O'$-$X'Y'Z'$. For the predicted depth under the virtual camera system, it has to take an affine transformation to recover the metric depth under the real camera system, i.e., $P_{A} = s\cdot (P_{A'} +T)$,  where $P=(x,y,d)^{T}$. The learning objective is defined as follows.
\begin{equation}
    L = \min_{\theta}\left | \mathcal{K}(\mathcal{G}({\bf I}, \theta)) - d^{\ast} \right |
\label{eq: total loss}
\end{equation}
where $\mathcal{K}(\cdot)$ is the affine transformation to recover the scaling and translation.
Through explicitly defining a virtual camera system and disentangling the affine transformation between the diverse real camera system and the virtual one, we simplify the objective of monocular depth prediction. The predicted depth will be invariant to various scales and translations. Therefore, it will be easier to generalize to diverse scenes by learning affine-invariant depth than metric depth. Besides, such learning objective can maintain more geometric information than that of learning relative depth.

In Table~\ref{table:loss_list}, several
losses
can disentangle the scaling and translation and enforce the model to learn the affine-invariant depth. The virtual normal loss (VNL) and surface normal loss~\cite{qi2018geonet, Yin2019enforcing} are constructed based on the normals, which are essentially invariable to scaling and translation. Furthermore, the scale-and-shift-invariant loss (SSIL)~\cite{lasinger2019towards} explicitly recovers the scaling and translation before minimizing the divergence to ground truth. Therefore, we take the high-order geometric
loss
and the SSIL to optimize the network. The overall loss function is illustrated as follows.
$\lambda$ is to balance the two terms, which is set $1$ in our experiments.

\begin{equation}
    \ell=L_{vn}(d, d^{\ast}) + \lambda L_{ssi}(d, d^{\ast})
\label{eq: total loss}
\end{equation}

\begin{table}[]
\centering
\caption{Illustration of different %
loss functions
}
\scalebox{0.6}{
\begin{tabular}{ r| c}
\toprule[1pt]
Loss
& Definition  \\ \hline \hline
\multicolumn{2}{c}{Metric Depth Loss} \\ \hline \hline
MSE      & $L_{mse} = \frac{1}{N}\sum_{i=1}^{N}\left ( d_{i} - d_{i}^{\ast } \right )^{2} $     \\ \hline
Silog~\cite{eigen2014depth}   & \begin{tabular}{@{}c@{}} $L_{si} = \frac{1}{N} \sum_{i=1}^{N}y_{i}^{2} - \frac{1}{N^{2}} (\sum_{i=1}^{N}y_{i})^{2}$ \\ $y_{i} = \log(\frac{d_{i}}{d_{i}^{\ast }})$ \end{tabular}   \\ \hline \hline
\multicolumn{2}{c}{Relative Depth Loss} \\ \hline \hline
Ranking~\cite{xian2018monocular} & $L_{rank} = \begin{cases}
 & \log(1 + \exp(d_{i}-d_{j})l_{ij}), l_{ij}=\pm 1\\
 &(d_{i} - d_{j})^{2}, l_{ij}= 0
\end{cases} $       \\ \hline \hline
\multicolumn{2}{c}{Affine-invariant Depth Loss} \\ \hline \hline
Scale-shift-invariant
~\cite{lasinger2019towards}  &\begin{tabular}{@{}c@{}} $L_{ssi}=\frac{1}{2N} \sum_{i=1}^{N} (\overrightarrow{\mathbf{d}_{i}^{\top}}\mathbf{h} - d_{i}^{\ast})^{2}$ \\ $\mathbf{h}=(\sum_{i=1}^{N}\overrightarrow{\mathbf{d}_{i}}\overrightarrow{\mathbf{d}_{i}^{\top}})^{-1}(\sum_{i=1}^{N}\overrightarrow{\mathbf{d}_{i}}\mathbf{d}_{i}^{\ast})$, $\overrightarrow{\mathbf{d}_{i}}=(d_{i}, 1)^{\top}$
\end{tabular}    \\ \hline
Virtual normal
\cite{Yin2019enforcing}     &\begin{tabular}{@{}c@{}} $L_{vn} = \frac{1}{N} (\sum_{i=0}^{N} \| {\mathbf{n}_{i}} -{\mathbf{n}_{i}^{\ast}}\|_{1}) $ \\ $n$ is the virtual normal  \end{tabular}  \\ \hline
Surface normal
~\cite{qi2018geonet}     & \begin{tabular}{@{}c@{}} $L_{sn} = \frac{1}{N} (\sum_{i=0}^{N} \| {\mathbf{n'}_{i}} -{\mathbf{n'}_{i}^{\ast}}\|_{1}) $ \\ $n'$ is the surface normal \end{tabular}   \\
 \toprule[1pt]
\end{tabular}}
\label{table:loss_list}
\end{table}

\subsection{Multi-curriculum Learning}
Most existing methods uniformly sample a sequence of mini-batches $[\mathbb{B}_{0},..., \mathbb{B}_{M}]$ from the whole dataset for training. However, as our \datasetshortname~has a wide range of scenes, experiments illustrate that such training paradigm cannot effectively optimize the network. We propose a multi-curriculum learning method to solve this problem.
We sort the training data by the increasing difficulty and sample a series of mini-batches that exhibit an increasing level of difficulty. Therefore, there are two problems that should be solved: $1$) how to construct the curriculum; $2$) how to yield a sequence of easy-to-hard mini-batches for the network. Pseudo-code for multi-curriculum algorithm is shown in Algorithm~\ref{alg: cl}.

\noindent\textbf{Constructing the curriculum.} Three parts of \datasetshortname, i.e., \textit{part-fore, part-in} and \textit{part-out}, are termed as $\mathbb{X} = \{\mathbb{D}_{j}\}^{P}_{j=0}$. Let $\mathbb{D}_{j} = \{(x_{ij}, y_{ij}) | i=0,...,N\}$ represents the $N$ data points of the part $j$, where $x_{ij}$ denotes a single data, $y_{ij}$ is the corresponding label. Previous monocular depth estimation
methods show that training on limited scenes are easy to converge, so we train three models, $\mathcal{G}_{j}$, separately on $3$ parts as teachers. The absolute relative error (Abs-Rel) is chosen as the \textit{scoring function} $\mathcal{F}(\cdot)$ to evaluate the difficulty of each training sample. If $\mathcal{F}(\mathcal{G}_{j}(x_{ij}), y_{ij}) > \mathcal{F}(\mathcal{G}_{j}(x_{(i+1)j}), y_{(i+1)j})$, then we define the data $(x_{ij}, y_{ij})$ is more difficult to learn. Finally, we sort $3$ parts according to the ascending Abs-Rel error and the ranked datasets are $\mathbb{C}_{j} = \{(x_{ij}, y_{ij}) | i=0,...,N\}$.

\noindent\textbf{Mini-batch sampling.}  The \textit{pacing function} $\mathcal{H}(\cdot)$ determines a sequence of subsets of the dataset so that the likelihood of the easier data will decrease in this sequence, i.e. $\{\mathbb{S}_{0j},\dots,\mathbb{S}_{Kj}\} \subseteq \mathbb{C}_{j}$, where $\mathbb{S}_{kj}$ represents the first $\mathcal{H}(k, j)$ elements of $\mathbb{C}_{j}$. From each subset $\mathbb{S}_{kj}$, a sequence of mini-batches $\{\mathbb{B}_{0j} ,..., \mathbb{B}_{Mj}|j=0,1,2\}$ are uniformly sampled. Here we utilize the stair-case function as the \textit{pacing function}, which is determined by the starting sampling percentage $p_{j}$, the current \textit{step} $k$, and the fixed \textit{step length} $I_{o}$ (the number of iterations in each step). In each \textit{step} $k$, there are $I_{o}$ iterations and the $\mathcal{H}(k, j)$ remains constant, thus the step $k = \left \lfloor \frac{iter}{I_{o}} \right\rfloor$, where $iter$ is the iteration index. $\mathcal{H}(k, j)$ is defined as follows.
\begin{equation}
    \mathcal{H} (k, j) = min(p_{j} \cdot k, 1)\cdot N_{j}
\label{eq: pacing function}
\end{equation}
where $N_{j}$ is the size of part $\mathbb{D}_{j}$.
\setlength{\textfloatsep}{0.15cm}
\begin{algorithm}[t]
 \begin{footnotesize}
  \SetAlgoLined
        \SetKwInOut{Input}{Input}
        \SetKwInOut{Output}{Output}
  \Input{scoring function $\mathcal{F}$, pacing function $\mathcal{H}$, dataset $\mathbb{X}$}
  \Output{mini-batches sequence $\{\mathbb{B}_{i} | i=0\dots M\}$.}
 train the model $\mathcal{G}_{j}$ on the data part $\mathbb{D}_{j}$ as the teacher \\
 sort each data part $\mathbb{D}_{j}$ with ascending difficulty according to $\mathcal{F}$, the ranked data is $\mathbb{C}_{j}$ \\

  \For{$k=0$ \KwTo $K$}
  {\vspace{0.1cm}
      \For{$i=0$ \KwTo $M$}
      {\vspace{0.1cm}
          \For{$j=0$ \KwTo $P$}
          {
              subset size $s_{kj} = \mathcal{H}(k, j)$ \\
              subset $\mathbb{S}_{kj} = \mathbb{C}_{j}[0,\dots,s_{kj}]$ \\
              uniformly sample batch $\mathbb{B}_{ij}$ from $\mathbb{S}_{kj}$ \\
          }
          concatenate $P$ batches sampled from different data parts together $\mathbb{B}_{i} = \{\mathbb{B}_{ij}\}_{j=0}^{P}$ \\
          append $\mathbb{B}_{i}$ to the mini-batches sequence
      }
  }
  \caption{multi-curriculum learning algorithm}
  \label{alg: cl}
 \end{footnotesize}
\end{algorithm}
\setlength{\floatsep}{0.15cm}

\section{Experiments}
In order to demonstrate the generalization and effectiveness of our method, we test our method quantitatively and qualitatively on several zero-shot datasets and compare it with other state-of-the-art
methods.

\noindent\textbf{Experiment setup.} We test on $8$ zero-shot datasets to illustrate the performance and generalization of our method, i.e., NYU~\cite{silberman2012indoor}, KITTI~\cite{geiger2013vision}, DIW~\cite{chen2016single}, ETH$3$D~\cite{schops2017multi}, ScanNet~\cite{dai2017scannet}, TUM-RGBD~\cite{sturm2012benchmark}, \datasetshortname-H-Realsense, and \datasetshortname-H-SIMU. The last two testing datasets are constructed by us to test the performance on scenes with foreground people. We use two different RGB-D sensors, Realsense and SIMU, to capture people in several indoor and outdoor scenes. \datasetshortname-H-Realsense contains $2329$ images, while \datasetshortname-H-SIMU has $8685$ images. %
We use the model applied in \cite{Yin2019enforcing} with the pre-trained ResNeXt-$50$~\cite{xie2017aggregated} backbone. The SGD is utilized for optimization with the initial learning rate of $0.0005$ for all layers. The learning rate is decayed every $5$K iterations with the ratio $0.9$. The batch size is set to $12$. Note that we evenly sample images from three data parts to constitute a batch. During the training, images are flipped horizontally, resized with the ratio from $0.5$ to $1.5$, and cropped with the size of $385 \times 385$. In the testing, we will resize, pad, and crop the image to keep a similar aspect ratio.

\noindent\textbf{Evaluation metrics.} We mainly take the absolute relative error (Abs-Rel) for evaluation except DIW, which is evaluated with the Weighted Human Disagreement Rate (WHDR)~\cite{xian2018monocular}. Besides, when evaluating the depth of foreground people, we follow the approach in~\cite{li2019learning} to take scale-invariant root mean squared error (Si-RMS) and Abs-Rel for evaluation. As our model can only predict the affine-invariant depth of the scene, we explicitly scale and translate the depth to recover the metric depth when evaluating the metric depth. The scaling and translation factors are obtained by the least-squares method.

\subsection{Comparison with State-of-the-art  Methods}

\begin{table*}[]
\centering
\setlength{\abovecaptionskip}{10pt}
\caption{The comparison with state-of-the-art  methods on five zero-shot datasets. Our method outperforms previous learning the relative depth or metric depth methods significantly. }
 \setlength{\tabcolsep}{4.2pt}
\scalebox{0.9}{
\begin{tabular}{l|l|l|l|llll}
\toprule[1pt]
\multirow{3}{*}{Method}& \multirow{3}{*}{\begin{tabular}{@{}c@{}} Training\\dataset \end{tabular}} & Backbone & \multicolumn{5}{c}{Testing on zero-shot datasets} \\ \cline{4-8} 
        &  &  &  DIW & NYU & KITTI & ETH3D & ScanNet \\
        &  &  &  WHDR & \multicolumn{4}{c}{Abs-Rel} \\ \hline \hline
\multicolumn{7}{c}{Learning Metric Depth + Single-scene Dataset} \\ \hline \hline
Yin~\etal~\cite{Yin2019enforcing} & NYU  & ResneXt-101 &$27.0$   &$\underline{10.8}$  &$35.1$  &$29.6$  &$13.66$ \\  
Alhashim~\etal~\cite{alhashim2018high} & NYU & DenseNet-169 &$26.8$   &$\underline{12.3}$   &$33.4$  &$34.5$   &$12.5$  \\ 
Yin~\etal~\cite{Yin2019enforcing} & KITTI & ResneXt-101 &$30.8$   &$26.7$  &$\underline{7.2}$  &$31.8$   &$23.5$   \\ 
Alhashim~\etal~\cite{alhashim2018high}& KITTI & DenseNet-169 &$30.9$  &$23.5$  &$\underline{9.3}$   &$32.1$  &$20.5$  \\ \hline \hline 
\multicolumn{7}{c}{Learning Relative Depth + Diverse-scene Dataset} \\ \hline \hline
Li and Snavely~\cite{li2018megadepth}&  MegaDepth &ResNet-50 &$24.6$  &$19.1$  &$19.3$  &$29.0$  &$18.3$ \\ 
Lasinger~\etal~\cite{lasinger2019towards} &MV + MegaDepth + RedWeb & ResNet-50 &$14.7$  &$19.1$  &$29.0$   &$23.3$ &$15.8$  \\ 
Chen~\etal~\cite{chen2016single}& DIW  & ResNeXt-50 &$\underline{11.5}$  &$16.7$ &$25.6$  &$25.7$  &$16.0$ \\ 
Xian~\etal~\cite{xian2018monocular}&  RedWeb   & ResNeXt-50 &$21.0$  &$26.6$   &$44.4$    &$39.0$  &$18.2$ \\  \hline \hline  
\multicolumn{7}{c}{Learning Affine-invariant Depth + Diverse-scene Dataset} \\ \hline \hline
Ours& \datasetshortname  & ResNeXt-50 &$\textbf{14.3}$ &$\textbf{11.7}$  &$\textbf{12.6}$ &$\textbf{22.5}$ &$\textbf{10.4}$ \\ \toprule[1pt] 
\end{tabular}}
\begin{tablenotes}
\footnotesize
\item '$\rule{0.3cm}{0.15mm}$' The method has trained the model on the corresponding dataset.
\end{tablenotes}
\label{table: zero-shot comparison}
\end{table*}

\noindent\textbf{Quantitative comparison on popular benchmarks.}
The quantitative comparison is illustrated in Table~\ref{table: zero-shot comparison}. Apart from Chen~\etal~\cite{chen2016single} and Xian~\etal~\cite{xian2018monocular}, whose performance is retrieved by re-implementing the ranking loss and training with our model, the performances of other methods are obtained by running their released codes and models. For all methods, we scale and translate the depth before evaluation. Those results whose models have been trained on the testing scene are marked with an underline.

Firstly, from Table~\ref{table: zero-shot comparison}, we can see that previous state-of-the-art  methods, which enforce the model to learn accurate metric depth, cannot generalize to other scenes. For example, the well-trained models of Yin~\etal~\cite{Yin2019enforcing} and Alhashim and Wonka~\cite{alhashim2018high} cannot perform well on other zero-shot scenes.

Secondly, although learning the relative depth methods can predict high-quality ordinal relations on diverse DIW dataset, i.e., one point being closer or further than another one, the discrepancy between the relative depth and the ground-truth metric depth is very large, see Abs-Rel on other datasets. Such high Abs-Rel results in these methods not being able to recover high-quality 3D shape of scenes, see Figure~\ref{fig:intro cmp} and Figure~\ref{fig:pcd cmp}.

By contrast, through enforcing the model to learn the affine-invariant depth and constructing a high-quality diverse dataset for training, our method can predict high-quality depths on various zero-shot scenes. Our method can outperform previous methods by up to $70\%$. Noticeably, on NYU, our performance is even on par with existing state-of-the-art methods which have trained on NYU (ours $11.7\%$ vs.\  Alhashim $12.3\%$).

\begin{figure*}[!bth]
\centering
\includegraphics[width=1\textwidth]{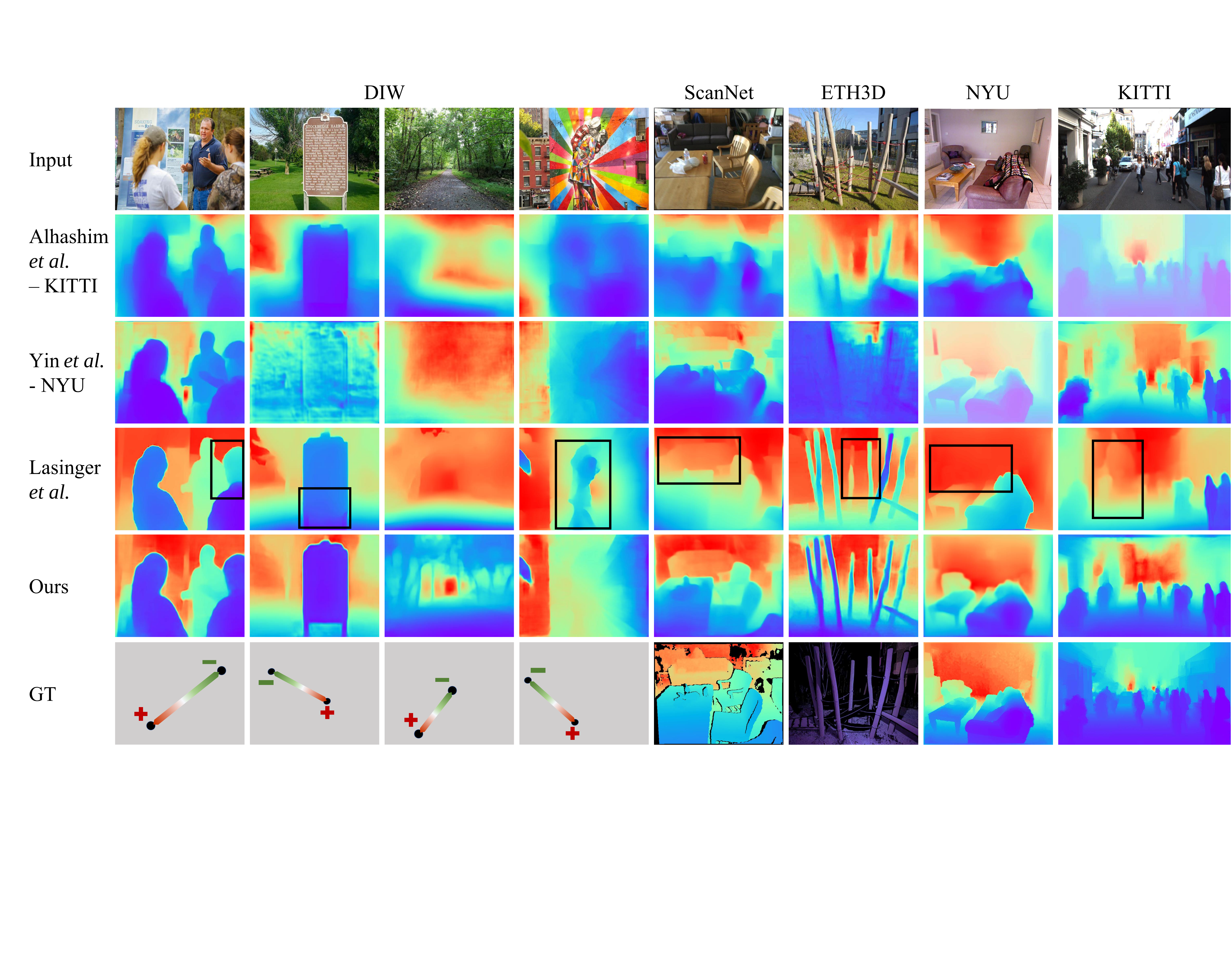}
\caption{Qualitative comparison with state-of-the-art  methods on zero-shot datasets. The transparent masks on images denote the method has been trained on the corresponding testing data. The black rectangles highlight the comparison regions. Our method not only predicts more accurate depth on diverse DIW, but also recovers better details on indoor and outdoor scenes, see marked regions on ScanNet, ETH3D, NYU, and KITTI. Note that ground truth of DIW only annotates the ordinal relation between two points.}
\label{fig:overall cmp}
\vspace{-1em}
\end{figure*}

\noindent\textbf{Qualitative comparison on zero-shot datasets.}
Figure~\ref{fig:overall cmp} illustrates the qualitative comparison on five zero-shot datasets. The transparent white masks denote the method has trained the model on the corresponding dataset. We can see those learning metric depth methods, Yin~\etal~\cite{Yin2019enforcing} and Alhashim and Wonka~\cite{alhashim2018high}, cannot work well on unseen scenes, while learning relative depth methods, see Lasinger~\etal, cannot recover high-quality depth map, especially for distant regions (see the marked regions on KITTI, NYU, and ScanNet) and regions with high texture difference (see marked head and colorful wall on DIW). On the DIW dataset, our method can predict more accurate depth on diverse DIW scenes, such as the forest and sign. Furthermore, on popular benchmarks, such as ScanNet, KITTI, and NYU, our method can also produce more accurate depth maps.

\noindent\textbf{Comparison of people.}
`People' is a significant foreground content for various applications. To our best knowledge, Li~\etal~\cite{li2019learning} are the first ones to focus on depth estimation for people. To promote the performance, they have to input the pre-computed depth of the background from two consecutive frames with structure from motion~\cite{schonberger2016pixelwise} and the mask of people regions to the network, see Li-IFCM and Li-IDCM in Table~\ref{table: people comparison}. Li-I denotes the method with a single image input for the network. By contrast, our method can also predict the high-quality depth for people with a still image. We make comparison on three datasets, i.e.,  TUM-RGBD, \datasetshortname-H-Realsense, and \datasetshortname-H-SIMU.

In Table~\ref{table: people comparison}, Si-env and Si-hum denote the Si-RMS errors of the background and people, respectively.  On TUM-RGBD, our method outperforms three configurations of Li~\etal~\cite{li2019learning} on foreground people up to $10\%$. Our overall performance, Si-RMS, is also much better. As Li-IDCM inputs the depth of the background, its Si-env error is lower than ours.

\datasetshortname-H-Realsense and \datasetshortname-H-SIMU have more scenes than TUM-RGBD. We compare our method with Li-I. It is clear that our method outperforms theirs significantly over all metrics with a still image.

\begin{figure}[!bth]
\centering
\includegraphics[width=0.45\textwidth]{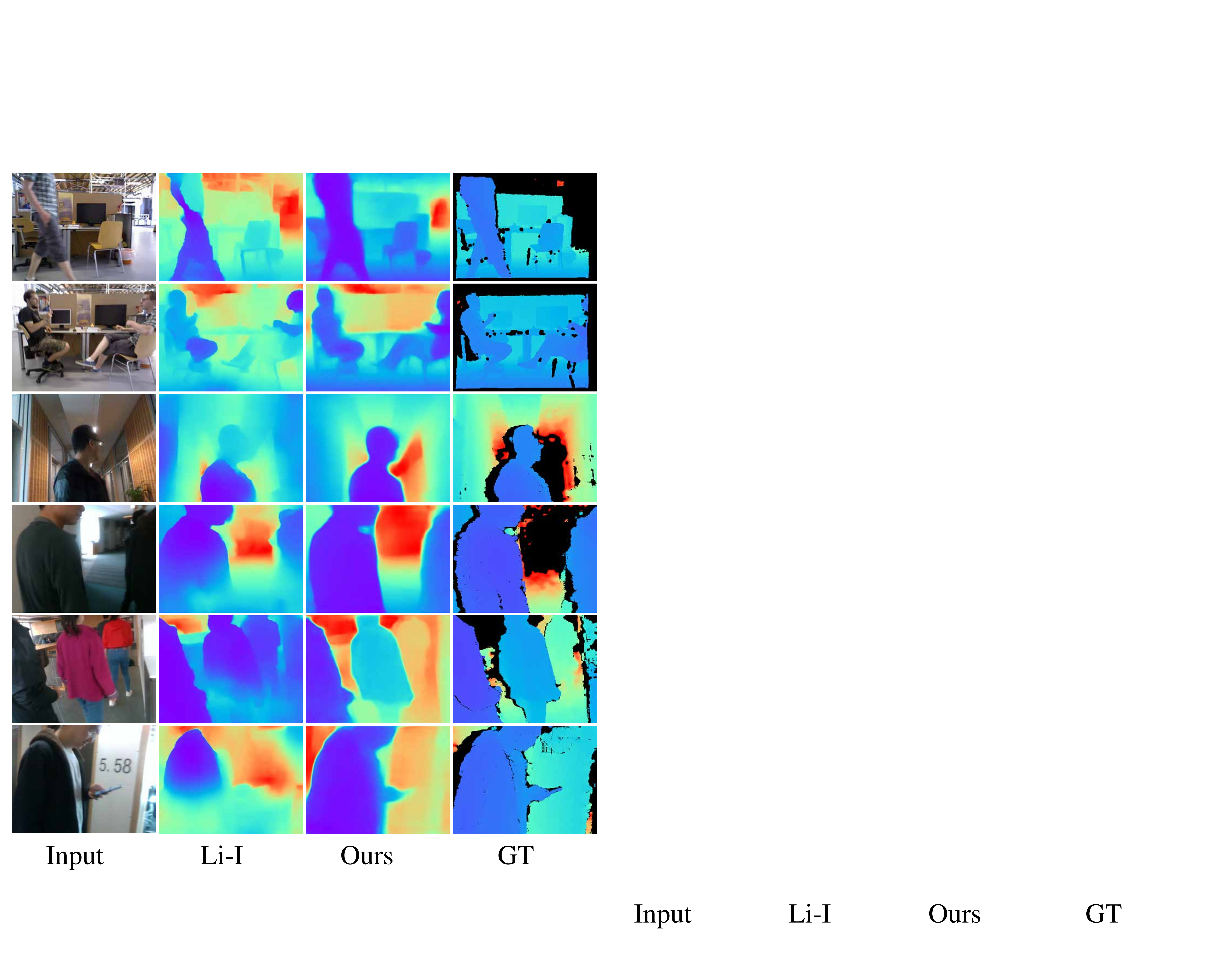}
\caption{Qualitative comparison of the foreground people. Our method and Li~\etal~\cite{li2019learning}-I have a single image input for the network. Our method can predict better depth on people and the background environment.}
\label{fig:people cmp}
\vspace{-2 em}
\end{figure}

Furthermore, we randomly select several images for qualitative comparison (see Figure~\ref{fig:people cmp}). It is clear that Li-I cannot perform well on the bottom part of people, distant people, and regions with significant texture difference, while our method can predict much better depths on both people and the background.

\begin{table}[]
\centering
\caption{The performance comparison of the foreground people on three zero-shot datasets. Our method can predict more accurate depth on foreground people over three datasets.}
\begin{threeparttable}
\scalebox{0.85}{
\begin{tabular}{llllll}
\toprule[1pt]
\multicolumn{1}{l|}{Method}  & \multicolumn{1}{l|}{Training} & \multicolumn{1}{c|}{Si-hum} & \multicolumn{1}{l|}{Si-env} & \multicolumn{1}{l|}{Si-RMS} & Abs-Rel \\ \hline \hline
\multicolumn{6}{c}{Testing on TUM-RGBD} \\ \hline \hline
\multicolumn{1}{l|}{Li-I}    & \multicolumn{1}{l|}{MC}&$0.294$  &$0.334$ &$0.318$ &$0.204$   \\
\multicolumn{1}{l|}{Li-IFCM\tnote{\dag}} & \multicolumn{1}{l|}{MC}&$0.302$ &$0.330$ &$0.316$ &$0.206$     \\
\multicolumn{1}{l|}{Li-IDCM\tnote{\dag}} & \multicolumn{1}{l|}{MC}&$0.293$ &$\textbf{0.238}$ &$0.272$ &$\textbf{0.147}$ \\
\multicolumn{1}{l|}{Ours} & \multicolumn{1}{l|}{\datasetshortname}&$\textbf{0.272}$ &$\underline{0.270}$ &$\textbf{0.272}$  &$\underline{0.192}$     \\ \hline \hline
\multicolumn{6}{c}{Testing on \datasetshortname-H-Realsense}\\ \hline \hline
\multicolumn{1}{l|}{Li-I}& \multicolumn{1}{l|}{MC}&$0.343$ &$0.305$ &$0.319$ &$0.264$     \\
\multicolumn{1}{l|}{Ours}    & \multicolumn{1}{l|}{\datasetshortname}&$\textbf{0.262}$  &$\textbf{0.241}$  &$\textbf{0.261}$  &$\textbf{0.186}$     \\ \hline\hline
\multicolumn{6}{c}{Testing on \datasetshortname-H-SIMU}\\ \hline \hline
\multicolumn{1}{l|}{Li-I}& \multicolumn{1}{l|}{MC}&$0.373$  &$0.466$  &$0.419$ &$0.391$     \\
\multicolumn{1}{l|}{Ours}    & \multicolumn{1}{l|}{\datasetshortname}&$\textbf{0.283}$   & $\textbf{0.335}$  &$\textbf{0.309}$    &$\textbf{0.218}$     \\
\toprule[1pt]
\end{tabular}}
\begin{tablenotes}
\footnotesize
\item[\dag] Input the mask of people and depth of background to the model.
\end{tablenotes}
\end{threeparttable}
\label{table: people comparison}
\end{table}

\subsection{Ablation Study}
In this section, we carry out several experiments to analyze the effectiveness of the proposed multi-curriculum learning  method, the effectiveness of different
loss functions
on the diverse data, the comparison of the reconstructed 3D point cloud among different methods, and the linear relations between the predicted affine-invariant depth and GT.

\begin{table}[]
\centering
\caption{The comparison of different training methods on $5$ zero-shot datasets and our \datasetshortname dataset. The proposed multi-curriculum learning method outperforms the baseline noticeably, while MCL-R can also promote the performance.}
\begin{threeparttable}
\scalebox{0.7}{
\begin{tabular}{l|lllll||ll}
\toprule[1pt]
\multicolumn{1}{c|}{\multirow{2}{*}{Method}} & DIW\tnote{\dag} & NYU\tnote{\dag} & KITTI\tnote{\dag} & ETH3D\tnote{\dag} & ScanNet\tnote{\dag} & \multicolumn{2}{c}{\datasetshortname}  \\ %
\multicolumn{1}{c|}{}  & \multicolumn{1}{c}{WHDR} & \multicolumn{4}{c||}{Abs-Rel}  &Abs-Rel &WHDR\\ \hline
Baseline &$14.5$   &$11.7$   &$17.9$  &$26.1$  & $11.2$   &$26.0$ & $16.4$     \\
MCL-R  &$15.0$  &$11.8$  &$15.8$  &$24.7$  &$11.0$  &$24.4$ & $15.9$       \\
MCL  &$\textbf{14.3}$ &$\textbf{11.7}$  &$\textbf{12.6}$ &$\textbf{22.5}$ &$\textbf{10.4}$  &$\textbf{20.6}$ &$\textbf{15.0}$  \\ \toprule[1pt]
\end{tabular}}
\begin{tablenotes}
\footnotesize
\item[\dag] Testing on zero-shot datasets.
\end{tablenotes}
\end{threeparttable}
\label{table: curriculum learning}
\end{table}

\begin{figure}[bth]
\centering
\includegraphics[width=0.4\textwidth]{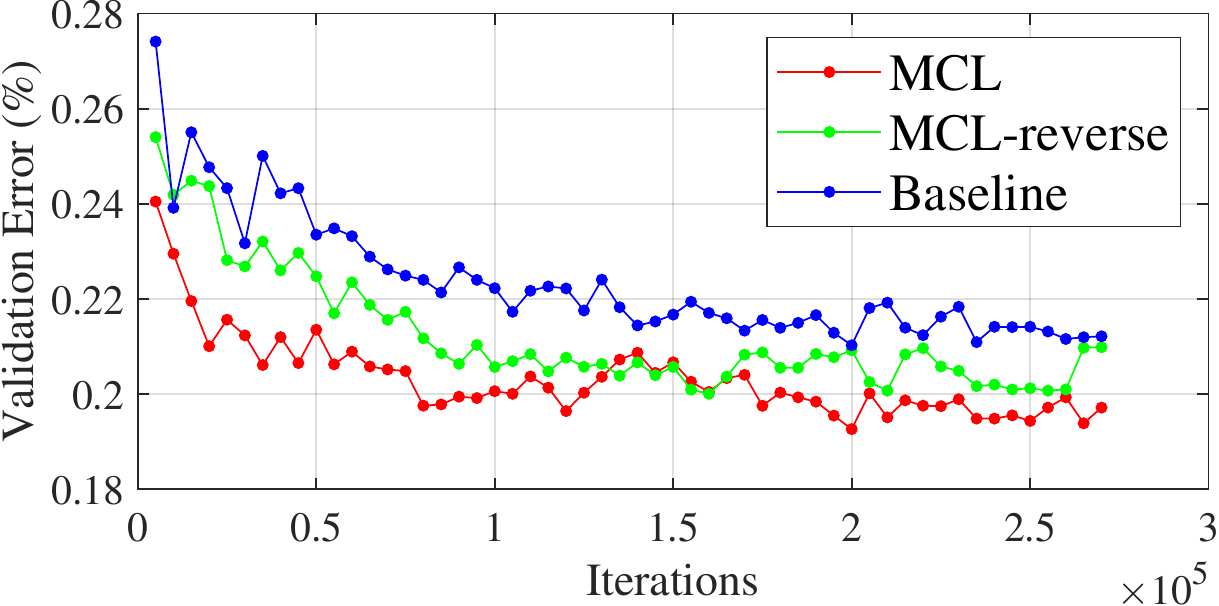}
\caption{Validation error during the training process. The validation error of the proposed
multi-curriculum learning
method is always lower than that of the MCL-R and baseline.}
\label{fig:effectiveness CL}
\end{figure}

\noindent\textbf{Effectiveness of multi-curriculum learning.}
To demonstrate the effectiveness of multi-curriculum learning method, we take three settings for the comparison: (1) sampling a sequence of mini-batches uniformly for training, termed Baseline; (2) using the reverse \textit{scoring function}, i.e., $\mathcal{F}^{'} = - \mathcal{F}$, thus the training samples are sorted in the descending order on difficulty and the harder examples are sampled more than easier ones, termed MCL-R; (3) using the proposed multi-curriculum learning method for training, termed MCL. We make comparisons on $5$ zero-shot datasets and our proposed \datasetshortname dataset. In Table \ref{table: curriculum learning}, it is clear that MCL outperforms the baseline by a large margin over all testing datasets. Although MCL-R can also promote the performance, it cannot equal MCL. Furthermore, we demonstrate the validation error along the training in Figure~\ref{fig:effectiveness CL}. It is clear that the validation error of  MCL is always lower than the baseline and  MCL-R over the whole training process. Therefore, the MCL method with an easy-to-hard curriculum can effectively train the model on diverse datasets.

\noindent\textbf{Effects of different losses.}
In this section, we analyze the effectiveness of various
loss functions
for
depth estimation on diverse datasets, including virtual normal loss (VNL), scale-shift-invariant loss (SSIL), Silog~\cite{eigen2014depth}, Ranking, and MSE. We sample $10$K images from each part of \datasetshortname separately for quick training then test the performances on $5$ zero-shot datasets. All the experiments take a multi-curriculum learning method. In Table~\ref{table: different constraints}, the VNL and SSIL outperform others over five zero-shot datasets significantly, which demonstrates the effectiveness of learning the affine-invariant depth on diverse datasets. By contrast, as MSE loss enforces the network to learn the accurate metric depth, it fails to generalize to unseen scenes, thus cannot perform well on zero-shot datasets. Although Ranking can make the model predict good relative depth on diverse DIW, Abs-Rel errors are very high on other datasets because it cannot enrich model with any geometric information. By contrast, as Silog considers the varying scale in the dataset, it performs a little better than Ranking and MSE.

\begin{table}[]
\centering
\caption{The effectiveness comparison of different losses on zero-shot datasets. VNL and SSIL outperform others noticeably. By contrast, the model supervised by MSE fails to generalize to diverse scenes, while Ranking can only enforce the model to learn the relative depth. Although Silog considers the varying scale in the dataset, its performance cannot equal VNL and SSIL.}
\scalebox{1}{
\small
\begin{tabular}{c |lllll}
\toprule[1pt]
\multicolumn{1}{c|}{\multirow{2}{*}{Loss}} &\multicolumn{5}{c}{Testing on zero-shot datasets}   \\ \cline{2-6}
\multicolumn{1}{c|}{}  & DIW & NYU & KITTI & ETH3D & ScanNet \\ \hline
VNL     &$\textbf{15.2}$ &$\textbf{12.2}$ &$21.0$ &$28.9$  &$\textbf{11.5}$ \\
SSIL  &$17.5$ &$16.5$ &$\textbf{16.3}$ &$\textbf{26.8}$  &$15.6$   \\
Silog   &$19.6$ &$20.8$ &$30.8$  &$29.4$   &$17.6$    \\
Ranking &$24.3$ &$23.4$ &$47.9$  &$39.5$  &$18.1$   \\
MSE     &$35.3$ &$33.2$ &$36.0$  &$30.2$  &$21.6$   \\ \toprule[1pt]
\end{tabular}}
\label{table: different constraints}
\end{table}

\begin{figure}[]
\centering
\includegraphics[width=.48\textwidth]{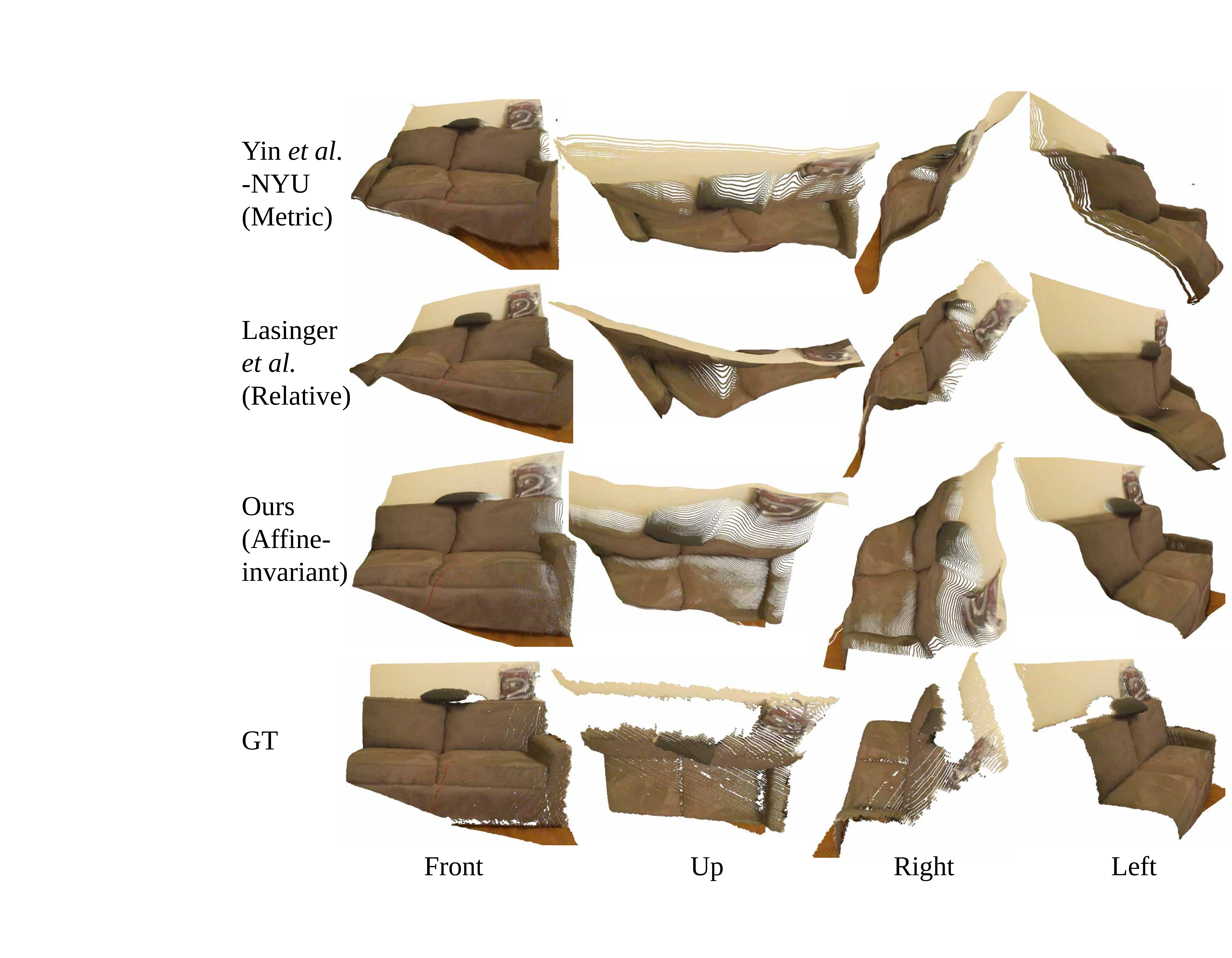}
\caption{Qualitative comparison of the reconstructed 3D point cloud from the predicted depth of a ScanNet image. Our method can clearly recover the shapes of the sofa and wall, while the shape of other methods distort noticeably. }.
\label{fig:pcd cmp}
\vspace{-1em}
\end{figure}

\noindent\textbf{Comparison of the recovered 3D shape.}
In order to further demonstrate learning affine-invariant depth can maintain the geometric information, we reconstruct the 3D point cloud  from the predicted depth of a random ScanNet image. We compare our methods with Lasinger~\etal~\cite{lasinger2019towards} and Yin-NYU~\cite{Yin2019enforcing}. We take four viewpoints for visual comparison, i.e., front, up, left, and right viewpoints. From Figure~\ref{fig:pcd cmp}, it is clear that our reconstructed point cloud can clearly represent the shape of the sofa and the wall from four views, while the sofa shapes of the other two methods are distorted noticeably and the wall is not flat.

\begin{figure}
\centering    %
\subfloat[] %
{
	\includegraphics[width=0.235\textwidth]{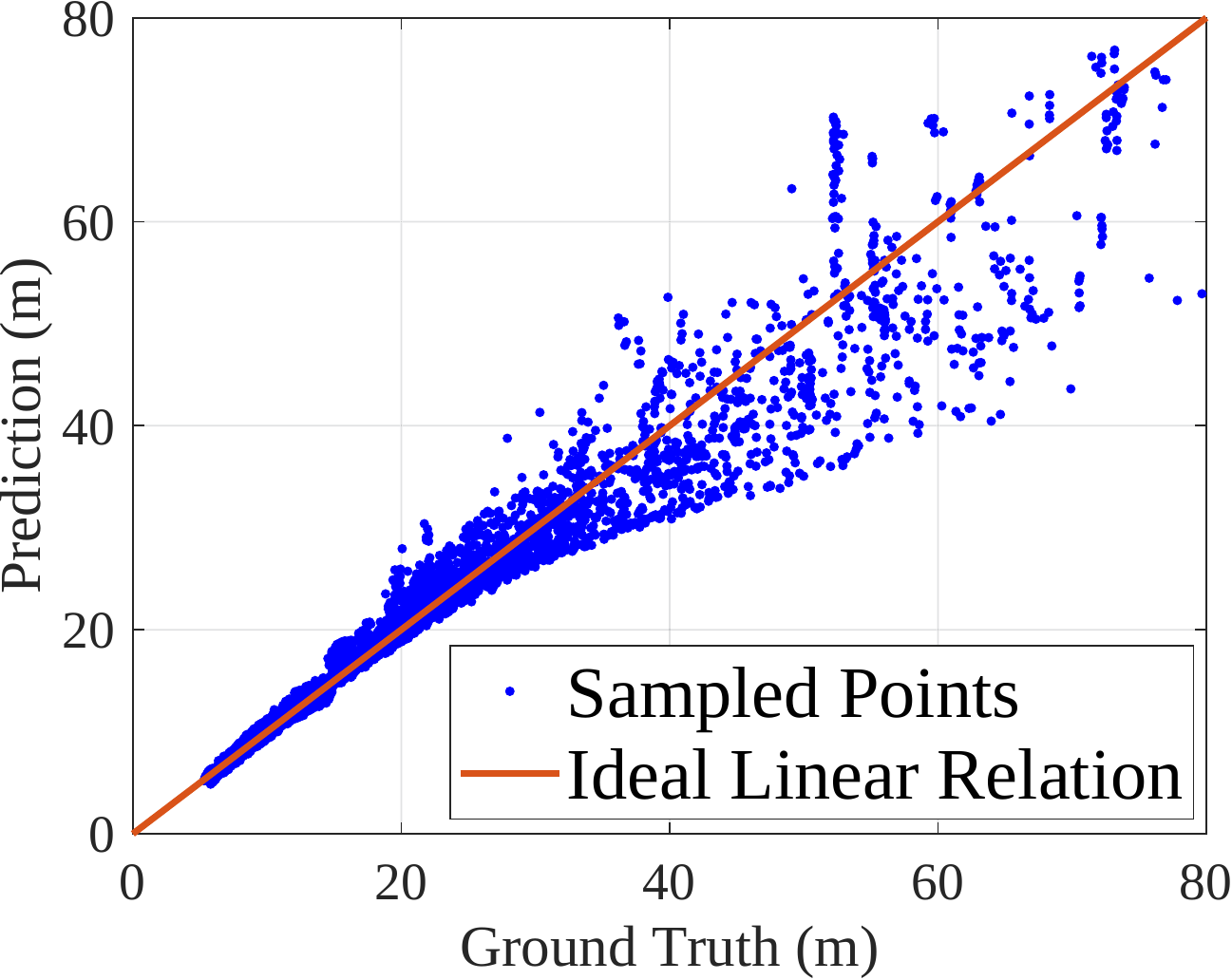}
	\label{fig:scaling shifting on kitti}
}
\subfloat[] %
{
	\includegraphics[width=0.23\textwidth]{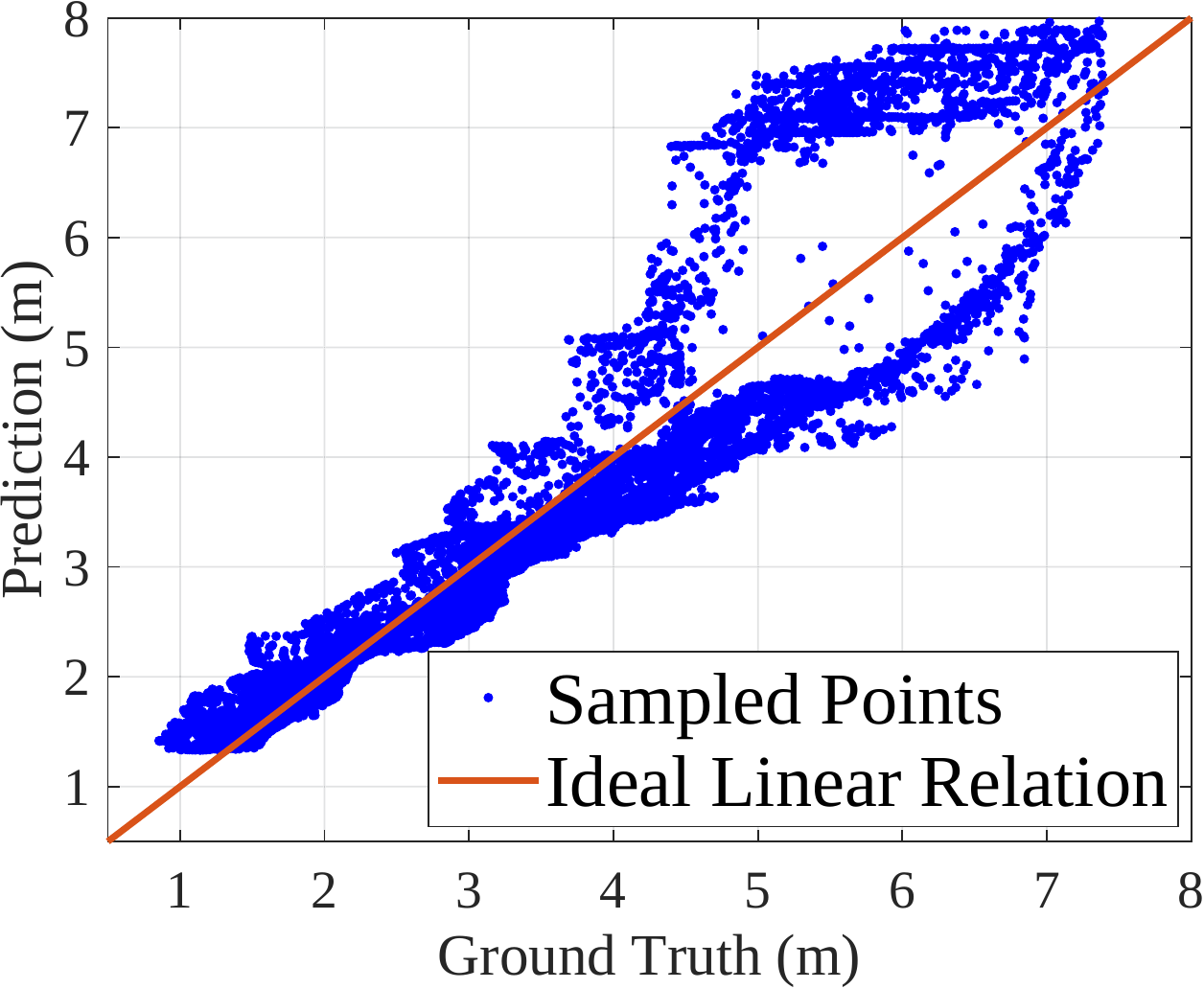}
	\label{fig:scaling shifting on nyu}
}
\caption{%
Testing the linear relation between the ground-truth and predicted depth. (a) Testing on KITTI. (b) Testing on NYU. Predicted depth has been scaled and translated for visualization. Blue points are the sampled points, while the red line is the ideal linear relation. There is roughly linear relation between the ground-truth and predicted depth. }
\label{fig:scaling and shifting}
\end{figure}

\noindent\textbf{Illustration of the affine transformation relation.}
To illustrate the affine transformation between the predicted affine-invariant depth and the ground-truth metric depth, we randomly select two images from KITTI and NYU respectively, and uniformly sample around $15$K points from each image. The predicted depth has been scaled and translated
for visualization.
In Figure~\ref{fig:scaling and shifting}, the red line is the ideal linear relation,
while the blue points are the sampled points. We can see the ground-truth depth and the predicted depth have a
roughly
linear relation. Note that as the precision of the sensor declines with the increase of  depth, as expected.
\section{Conclusions}
We
have attempted to
solve the generalization issue of monocular depth estimation,
at the same time maintaining as much
geometric information  as possible. Firstly we construct a large-scale and highly diverse RGB-D dataset. Compared with previous diverse datasets, which only have sparse depth ordinal annotations, our dataset is annotated with
dense and high-quality depth. Besides, we have proposed methods to learn the affine-invariant depth on our \datasetshortname dataset, which can ensure both %
good
generalization and high-quality geometric shape reconstruction from the depth. Furthermore, we propose a multi-curriculum learning method to train the model effectively on this diverse dataset. Experimental results on $8$ unseen datasets have shown the
usefulness
of our dataset and  method.

{\small
\bibliographystyle{ieee_fullname}
\bibliography{egbib}
}

\section{Supplementary}

\subsection{Details of the Datasets Used for Testing}
We conduct experiments on $7$ zero-shot datasets, including NYU~\cite{silberman2012indoor}, KITTI~\cite{geiger2013vision}, DIW~\cite{chen2016single}, ScanNet~\cite{dai2017scannet}, Eth3D~\cite{schops2017multi}, \datasetshortname-H-Realsense, and \datasetshortname-H-SIMU, to demonstrate the generalization of our proposed learning affine-invariant depth method in diverse scenes. The details of these datasets are illustrated as follows.

\begin{figure*}[bth]
\centering
\includegraphics[width=0.85\textwidth]{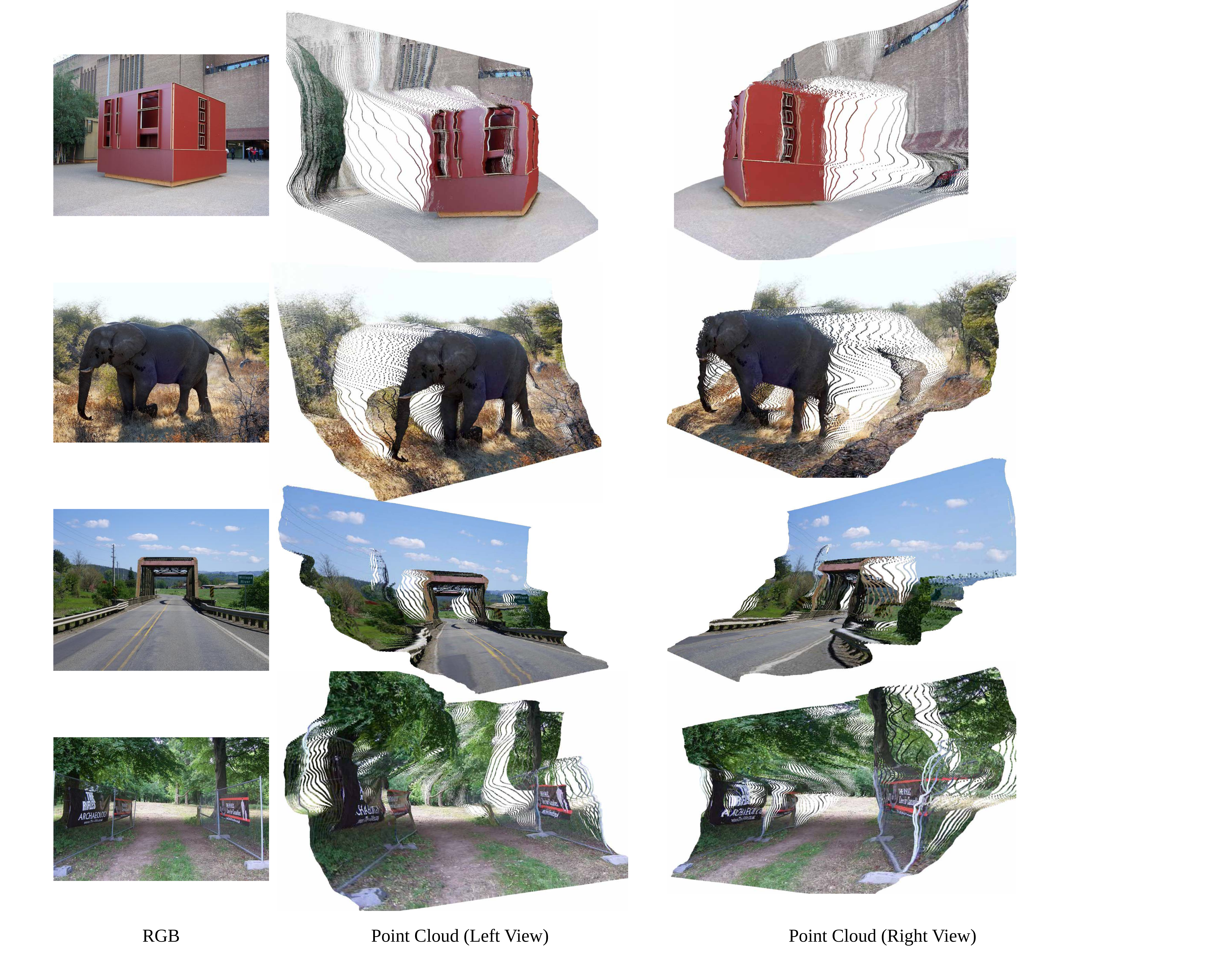}
\caption{3D point cloud. We randomly select several DIW images and reconstruct the 3D point cloud.}
\label{fig:pcd_1}
\vspace{-2em}
\end{figure*}

\begin{figure*}[bth]
\centering
\includegraphics[width=0.86\textwidth]{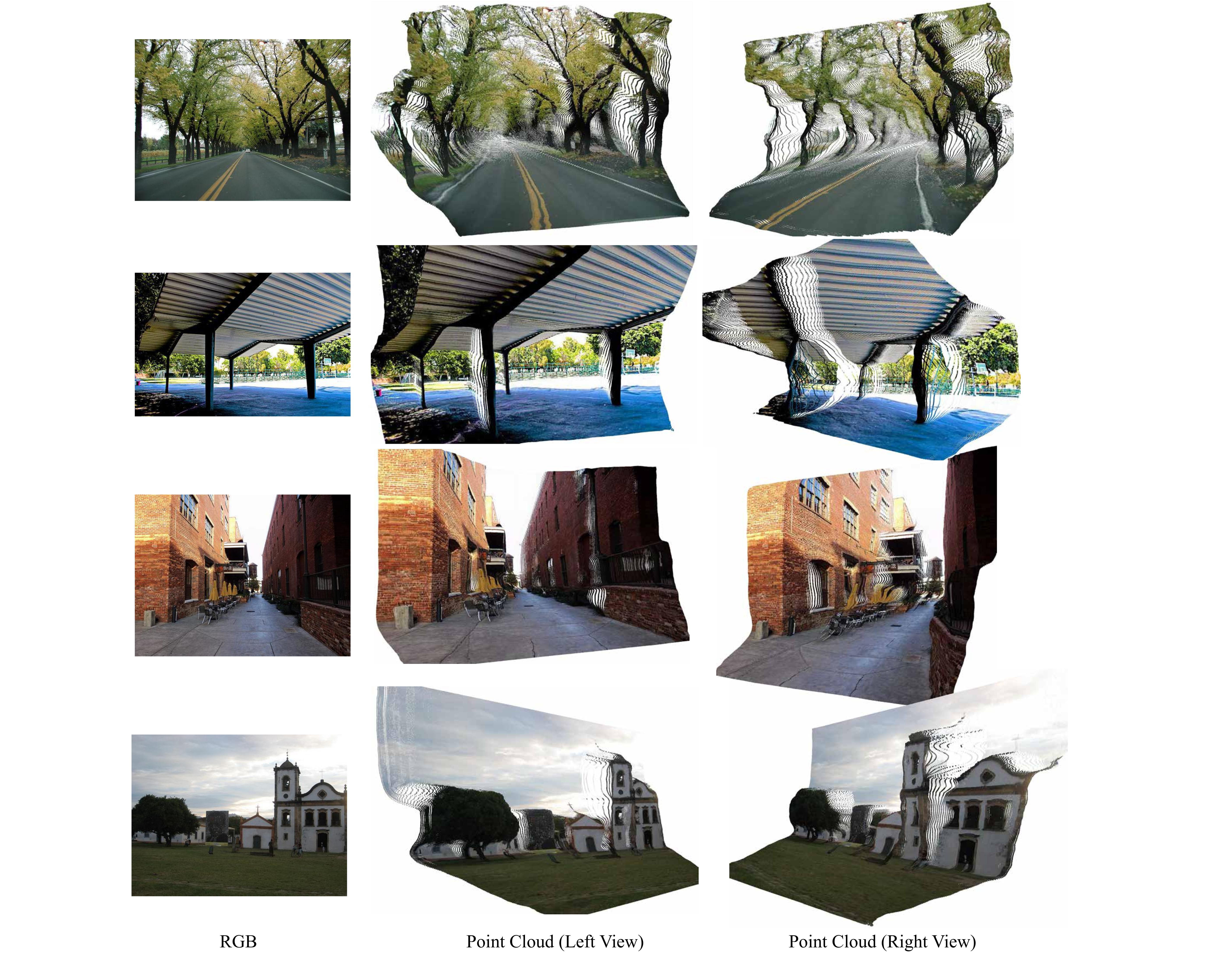}
\caption{3D point cloud. We randomly select several DIW images and reconstruct the 3D point cloud from the predicted affine-invariant depth. It is clear that the predicted depth can recover high-quality 3D shape information.}
\label{fig:pcd_2}
\end{figure*}

(1) \textbf{NYU.} We test on the testing set of NYU (Eigen split), which contains $654$ images.

(2) \textbf{KITTI.} We test on the testing set of KITTI (Eigen split), which contains $697$ images.

(3) \textbf{DIW.} We test on the testing set of DIW, which contains $74441$ images.

(4) \textbf{ETH3D.} We sample $105$ images from several scenes of ETH3D dataset for testing.

(5) \textbf{ScanNet.} We sample $2692$ images with 155 scenes from the validation set of ScanNet for testing.

(6) \textbf{\datasetshortname-H-Realsense.} We use the RealSense RGB-D camera to capture around $2329$ images in indoor and outdoor scenes. Note that we only test the contents between $0$-$5$m range.

(7) \textbf{\datasetshortname-H-SIMU.} We use the SIMU RGB-D camera to capture around $8685$ images in indoor scenes. The depth range is $0$-$10$m.

\vspace{-1.5em}

\subsection{Details of Evaluation Protocol}
Before evaluating the Abs-Rel error and Si-RMS error, we will scale and translate the predicted affine-invariant depth to recover the metric depth. The predicted depth will be in the same scale to the ground truth depth. The scaling and translation factors are obtained by the least-squares method.

\subsection{Illustrations of Predicted Affine-Invariant Depth on Diverse Images}
\begin{figure*}[!bth]
\centering
\includegraphics[width=0.9\textwidth]{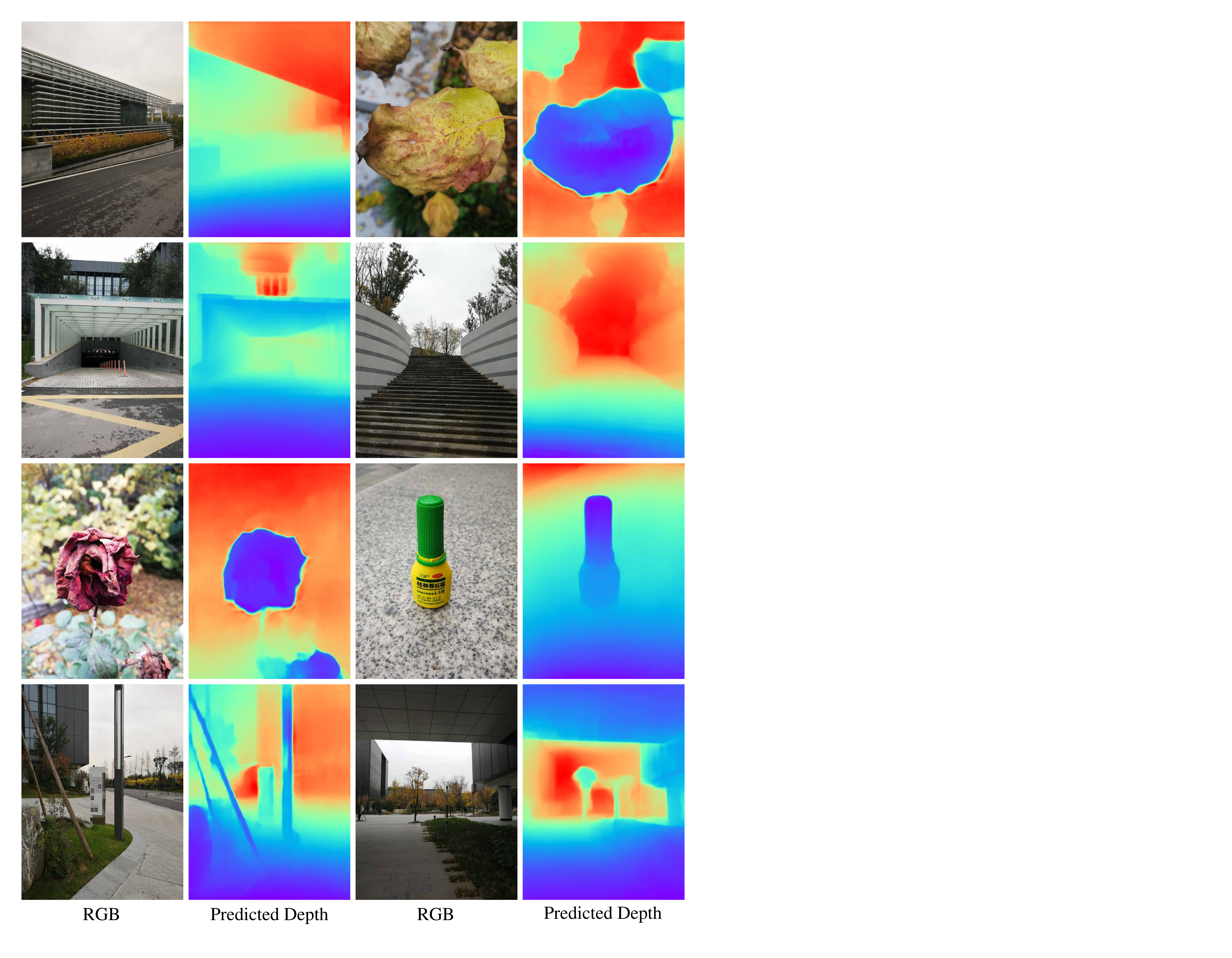}
\caption{Performance illustration. We randomly download several high-resolution images for testing. Our method performs well on diverse scenes.}
\label{fig:sm_3}
\end{figure*}

In order to further demonstrate the generalization of our proposed learning affine-variant depth, we randomly download several high-resolution images online for testing, see Figure~\ref{fig:sm_3}. We also show more results of DIW dataset. It is clear that our method can predict high-quality depth map for various contents in indoor and outdoor scenes.

\subsection{Illustrations of Reconstructed 3D Point Cloud}
In order to further demonstrate the quality of predicted affine-invariant depth, we reconstruct 3D point cloud from the depth map, see Figure~\ref{fig:pcd_2} and Figure~\ref{fig:pcd_1}. We randomly select several images from DIW for testing. Note that all these images have never been seen by our model during training. It is clear that the predicted depth can maintain high quality geometric shape information.

\subsection{Illustration of \datasetshortname Dataset}
\begin{figure*}[!bth]
\centering
\includegraphics[width=1\textwidth]{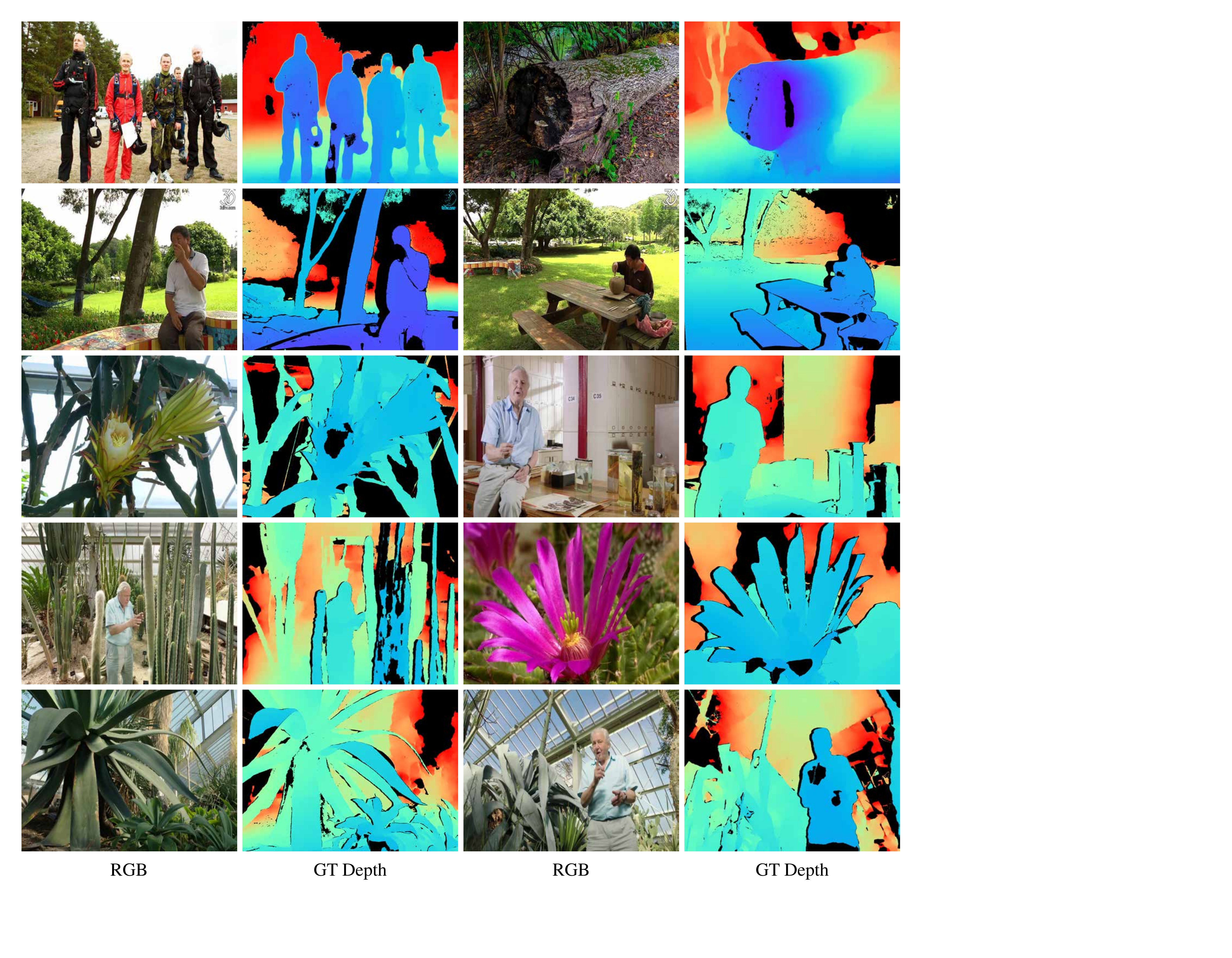}
\caption{\datasetshortname Dataset Illustration. We illustrate more RGB-D pairs of our proposed \datasetshortname dataset. We can see that our dataset contains a wide range of scenes. Note that the black regions in the GT depth maps will be masked out during training.}
\label{fig:sm_7}
\end{figure*}

 We illustrate more rgb and the corresponding ground truth depth maps in Figure \ref{fig:sm_7}. It is clear that our constructed \datasetshortname dataset comprehends a wide range of scenes.

\begin{figure*}[!bth]
\centering
\includegraphics[width=0.95\textwidth]{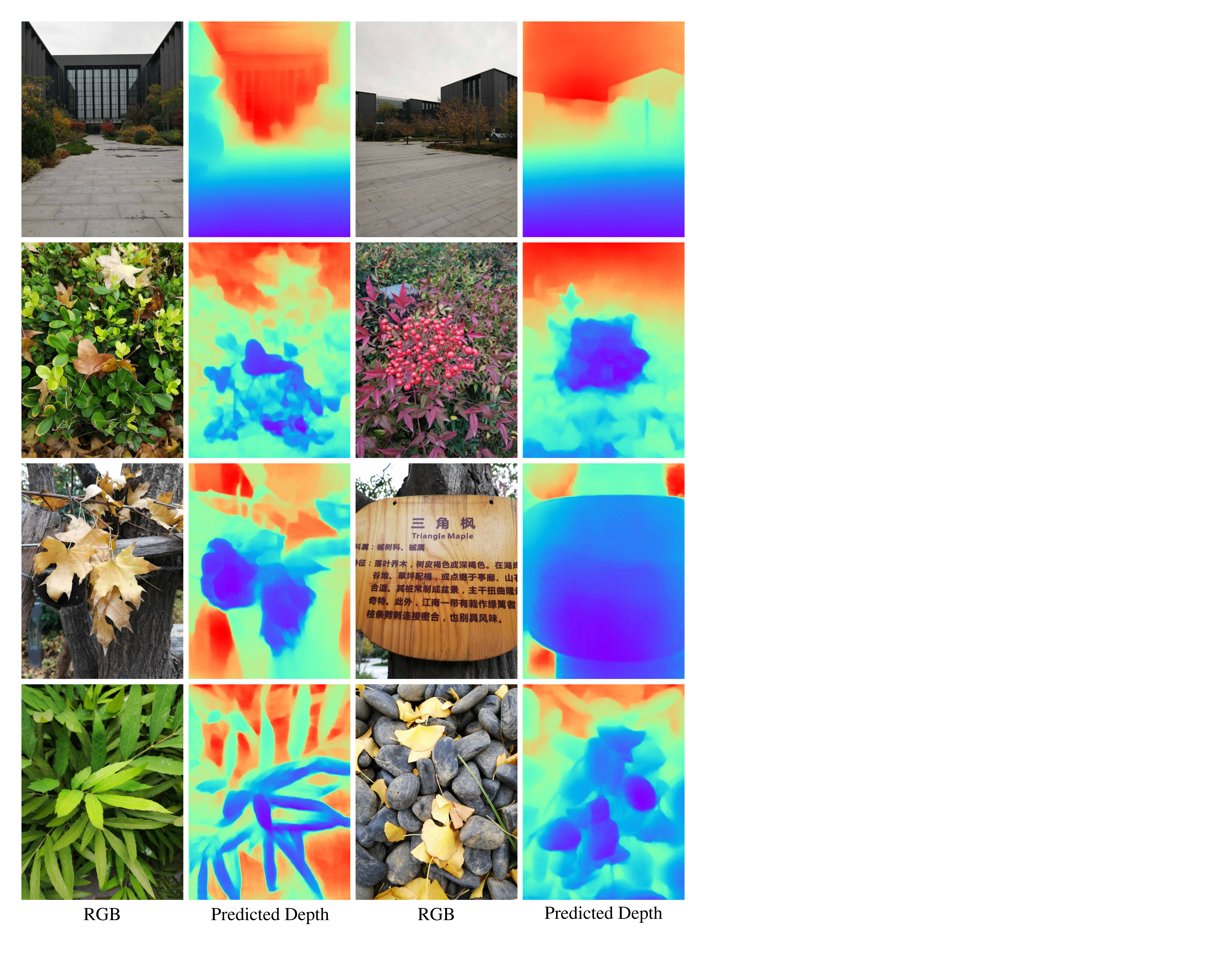}
\label{fig:sm_1}
\end{figure*}

\begin{figure*}[!bth]
\centering
\includegraphics[width=0.95\textwidth]{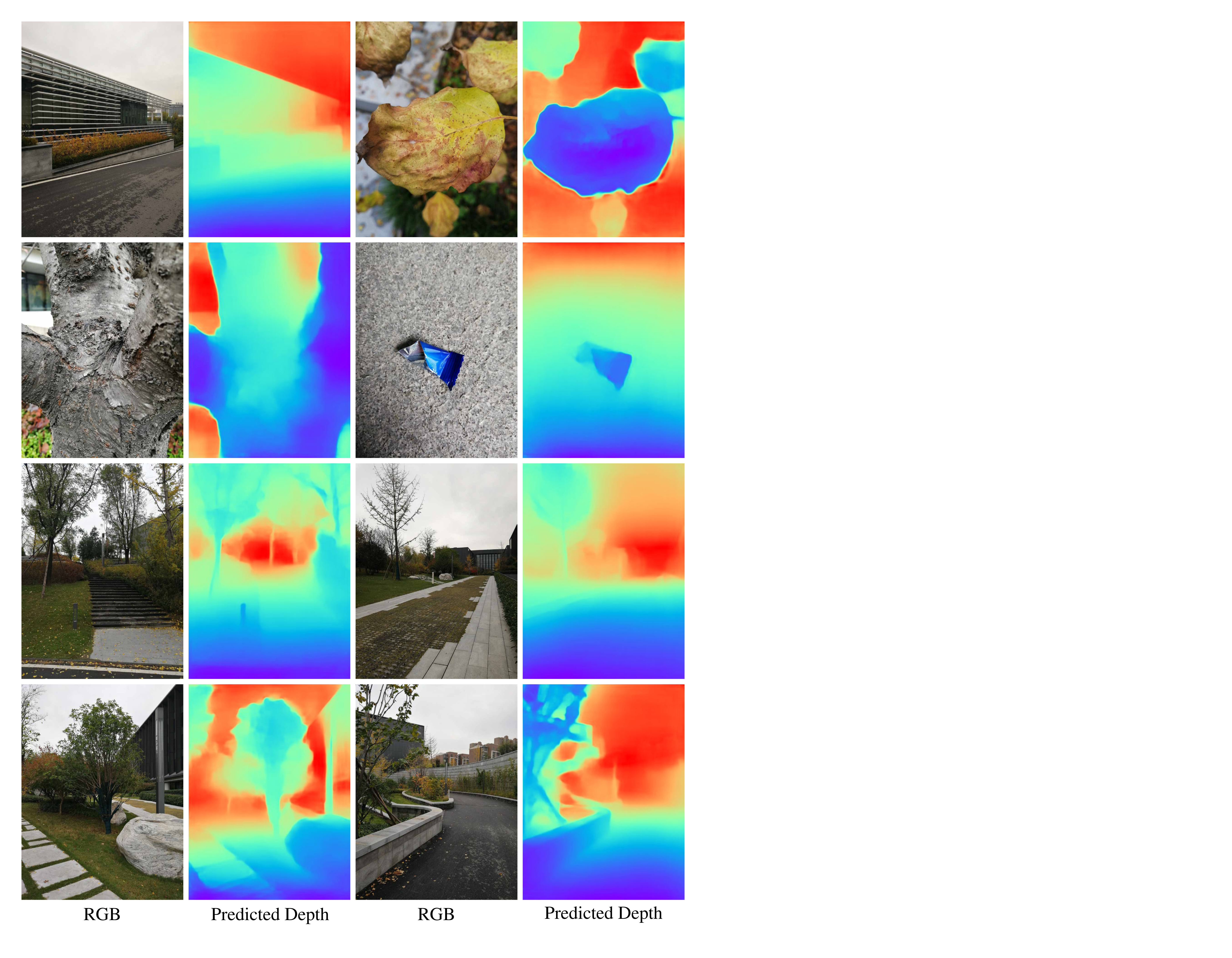}
\label{fig:sm_2}
\end{figure*}

\begin{figure*}[!bth]
\centering
\includegraphics[width=0.75\textwidth]{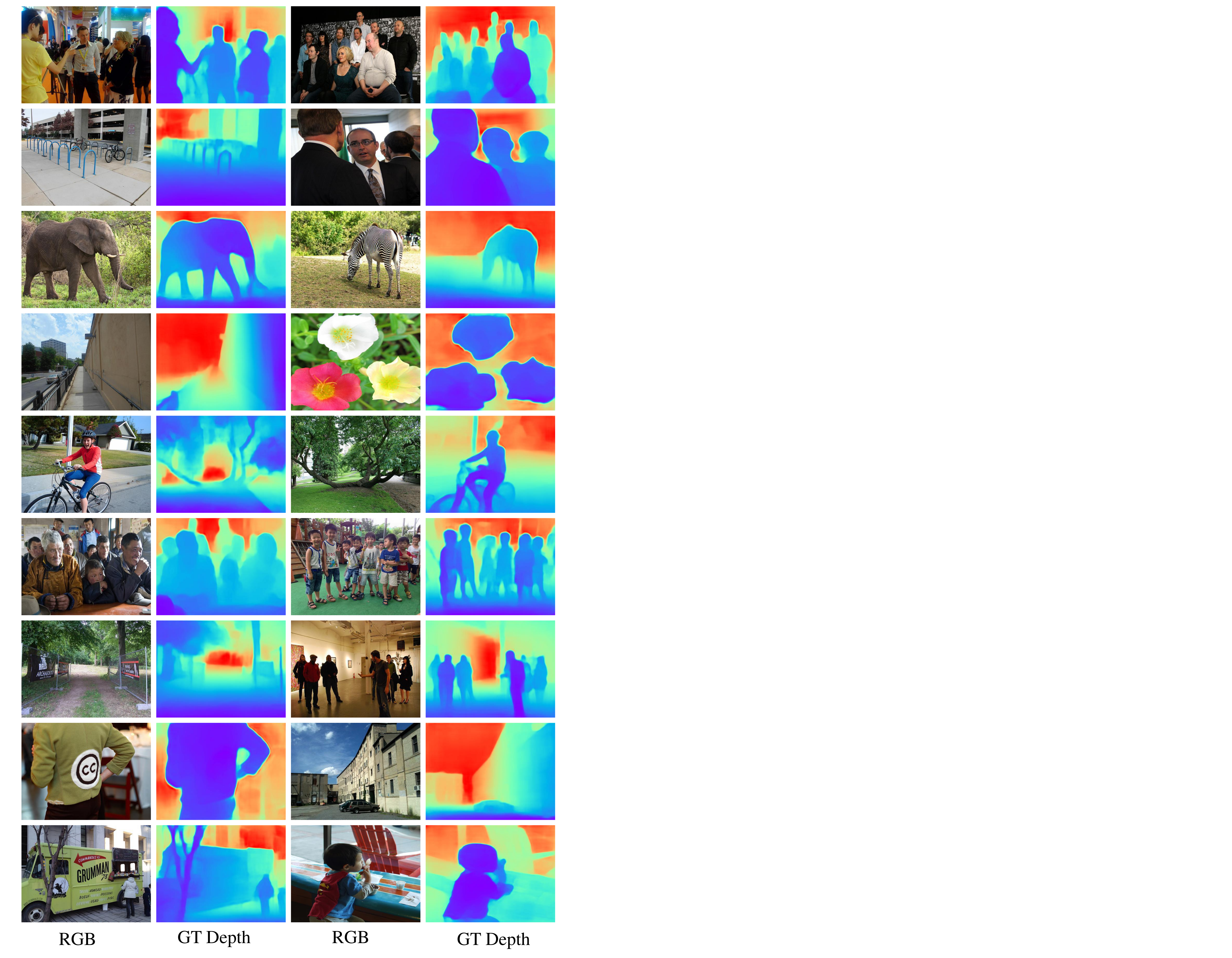}
\label{fig:sm_9}
\end{figure*}

\begin{figure*}[!bth]
\centering
\includegraphics[width=1\textwidth]{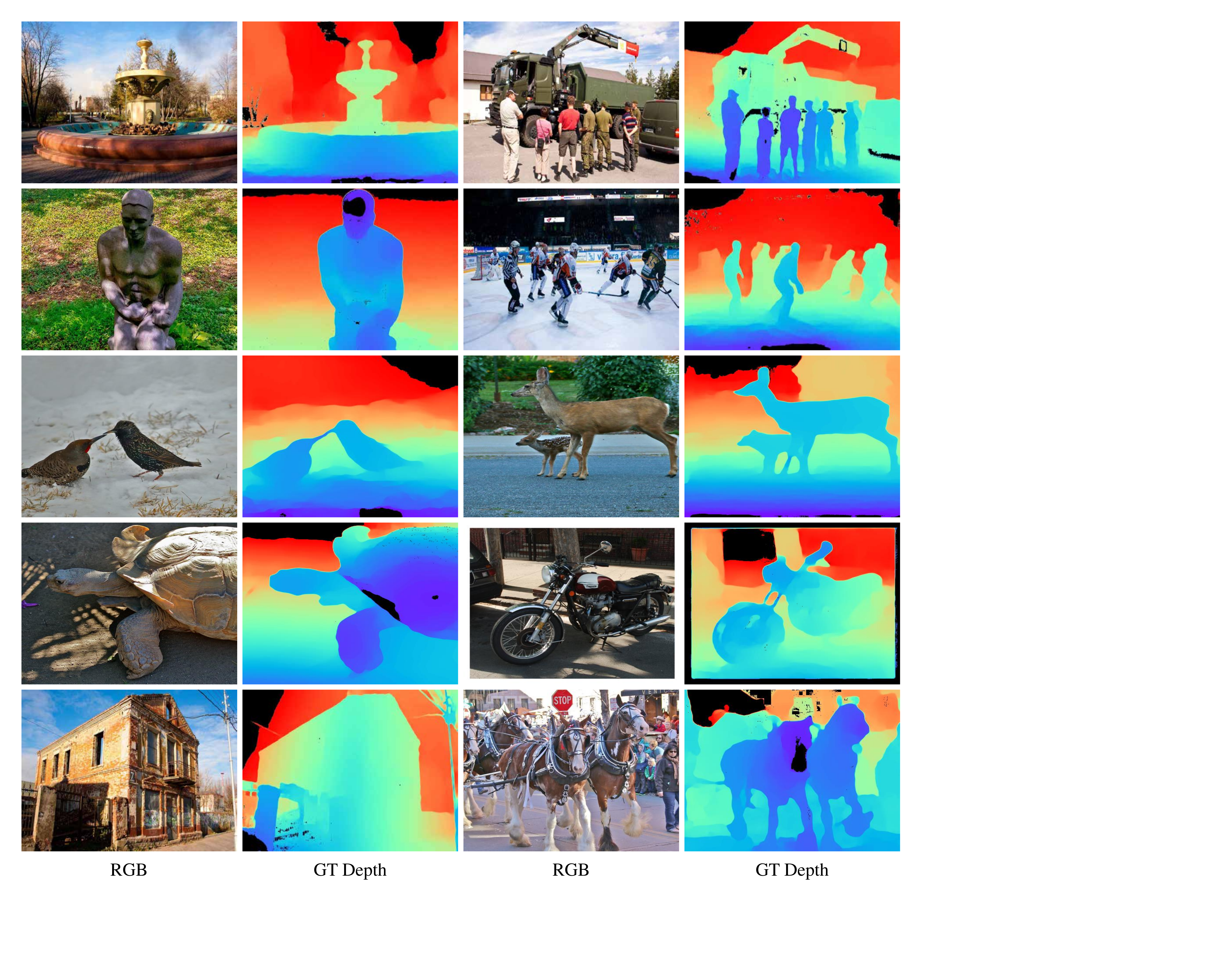}
\label{fig:sm_6}
\end{figure*}

\end{document}